\documentclass[10pt,twocolumn,letterpaper]{article}

\usepackage{iccv}
\usepackage{times}
\usepackage{epsfig}
\usepackage{graphicx}
\usepackage{amsmath}
\usepackage{amssymb}
\usepackage{pifont}
\usepackage{multirow}
\usepackage{booktabs}
\usepackage{mathtools}
\usepackage{comment}
\usepackage{subcaption}
\usepackage{enumitem}
\usepackage{listings}
\usepackage[dvipsnames]{xcolor}
\usepackage[pagebackref=true,breaklinks=true,letterpaper=true,colorlinks]{hyperref}

\iccvfinalcopy %

\ificcvfinal\fi

\begin{document}

\title{ADAPT: Efficient Multi-Agent Trajectory Prediction with Adaptation}

\author{Görkay Aydemir$^{1}$ \quad Adil Kaan Akan$^{1, 2}$ \quad Fatma Güney$^{1, 2}$\\
$^{1}$KUIS AI Center \quad\quad $^{2}$Department of Computer Engineering, Koç University\\
{\tt\small {gorkayaydemir}@gmail.com} \quad {\tt\small {kakan20}@ku.edu.tr} \quad {\tt\small {fguney}@ku.edu.tr}}

\newcommand{\Perp}{\perp\!\!\! \perp}
\newcommand{\bK}{\mathbf{K}}
\newcommand{\bX}{\mathbf{X}}
\newcommand{\bY}{\mathbf{Y}}
\newcommand{\bk}{\mathbf{k}}
\newcommand{\bx}{\mathbf{x}}
\newcommand{\by}{\mathbf{y}}
\newcommand{\bhy}{\hat{\mathbf{y}}}
\newcommand{\bty}{\tilde{\mathbf{y}}}
\newcommand{\bG}{\mathbf{G}}
\newcommand{\bI}{\mathbf{I}}
\newcommand{\bg}{\mathbf{g}}
\newcommand{\bS}{\mathbf{S}}
\newcommand{\bs}{\mathbf{s}}
\newcommand{\bM}{\mathbf{M}}
\newcommand{\bw}{\mathbf{w}}
\newcommand{\eye}{\mathbf{I}}
\newcommand{\bU}{\mathbf{U}}
\newcommand{\bV}{\mathbf{V}}
\newcommand{\bW}{\mathbf{W}}
\newcommand{\bn}{\mathbf{n}}
\newcommand{\bv}{\mathbf{v}}
\newcommand{\bwv}{\mathbf{wv}}
\newcommand{\bq}{\mathbf{q}}
\newcommand{\bR}{\mathbf{R}}
\newcommand{\bi}{\mathbf{i}}
\newcommand{\bj}{\mathbf{j}}
\newcommand{\bp}{\mathbf{p}}
\newcommand{\bt}{\mathbf{t}}
\newcommand{\bJ}{\mathbf{J}}
\newcommand{\bu}{\mathbf{u}}
\newcommand{\bB}{\mathbf{B}}
\newcommand{\bD}{\mathbf{D}}
\newcommand{\bz}{\mathbf{z}}
\newcommand{\bP}{\mathbf{P}}
\newcommand{\bC}{\mathbf{C}}
\newcommand{\bA}{\mathbf{A}}
\newcommand{\bZ}{\mathbf{Z}}
\newcommand{\bff}{\mathbf{f}}
\newcommand{\bF}{\mathbf{F}}
\newcommand{\bo}{\mathbf{o}}
\newcommand{\bO}{\mathbf{O}}
\newcommand{\bc}{\mathbf{c}}
\newcommand{\bm}{\mathbf{m}}
\newcommand{\bT}{\mathbf{T}}
\newcommand{\bQ}{\mathbf{Q}}
\newcommand{\bL}{\mathbf{L}}
\newcommand{\bl}{\mathbf{l}}
\newcommand{\ba}{\mathbf{a}}
\newcommand{\bE}{\mathbf{E}}
\newcommand{\bH}{\mathbf{H}}
\newcommand{\bd}{\mathbf{d}}
\newcommand{\br}{\mathbf{r}}
\newcommand{\be}{\mathbf{e}}
\newcommand{\bb}{\mathbf{b}}
\newcommand{\bh}{\mathbf{h}}
\newcommand{\bhh}{\hat{\mathbf{h}}}
\newcommand{\btheta}{\boldsymbol{\theta}}
\newcommand{\bTheta}{\boldsymbol{\Theta}}
\newcommand{\bpi}{\boldsymbol{\pi}}
\newcommand{\bphi}{\boldsymbol{\phi}}
\newcommand{\bpsi}{\boldsymbol{\psi}}
\newcommand{\bPhi}{\boldsymbol{\Phi}}
\newcommand{\bmu}{\boldsymbol{\mu}}
\newcommand{\bsigma}{\boldsymbol{\sigma}}
\newcommand{\bSigma}{\boldsymbol{\Sigma}}
\newcommand{\bGamma}{\boldsymbol{\Gamma}}
\newcommand{\bbeta}{\boldsymbol{\beta}}
\newcommand{\bomega}{\boldsymbol{\omega}}
\newcommand{\blambda}{\boldsymbol{\lambda}}
\newcommand{\bLambda}{\boldsymbol{\Lambda}}
\newcommand{\bkappa}{\boldsymbol{\kappa}}
\newcommand{\btau}{\boldsymbol{\tau}}
\newcommand{\balpha}{\boldsymbol{\alpha}}
\newcommand{\nR}{\mathbb{R}}
\newcommand{\nN}{\mathbb{N}}
\newcommand{\nL}{\mathbb{L}}
\newcommand{\cN}{\mathcal{N}}
\newcommand{\cA}{\mathcal{A}}
\newcommand{\cM}{\mathcal{M}}
\newcommand{\cR}{\mathcal{R}}
\newcommand{\cB}{\mathcal{B}}
\newcommand{\cG}{\mathcal{G}}
\newcommand{\cL}{\mathcal{L}}
\newcommand{\cH}{\mathcal{H}}
\newcommand{\cS}{\mathcal{S}}
\newcommand{\cT}{\mathcal{T}}
\newcommand{\cO}{\mathcal{O}}
\newcommand{\cC}{\mathcal{C}}
\newcommand{\cP}{\mathcal{P}}
\newcommand{\cE}{\mathcal{E}}
\newcommand{\cI}{\mathcal{I}}
\newcommand{\cF}{\mathcal{F}}
\newcommand{\cK}{\mathcal{K}}
\newcommand{\cV}{\mathcal{V}}
\newcommand{\cY}{\mathcal{Y}}
\newcommand{\cX}{\mathcal{X}}
\def\bgamma{\boldsymbol\gamma}

\newcommand{\specialcell}[2][c]{%
  \begin{tabular}[#1]{@{}c@{}}#2\end{tabular}}

\newcommand{\figref}[1]{\Fig~\ref{#1}}
\newcommand{\secref}[1]{Section~\ref{#1}}
\newcommand{\algref}[1]{Algorithm~\ref{#1}}
\newcommand{\eqnref}[1]{Eq.~\ref{#1}}
\newcommand{\tabref}[1]{Table~\ref{#1}}

\newcommand{\rulesep}{\unskip\ \vrule\ }

\newcommand{\KLD}[2]{D_{\mathrm{KL}} \Big(#1 \mid\mid #2 \Big)}

\renewcommand{\b}{\ensuremath{\mathbf}}

\def\mc{\mathcal}
\def\mb{\mathbf}

\newcommand{\T}{^{\raisemath{-1pt}{\mathsf{T}}}}

\makeatletter
\DeclareRobustCommand\onedot{\futurelet\@let@token\@onedot}
\def\@onedot{\ifx\@let@token.\else.\null\fi\xspace}
\def\eg{e.g\onedot} \def\Eg{E.g\onedot}
\def\ie{i.e\onedot} \def\Ie{I.e\onedot}
\def\cf{cf\onedot} \def\Cf{Cf\onedot}
\def\etc{etc\onedot} \def\vs{vs\onedot}
\def\wrt{wrt\onedot}
\def\dof{d.o.f\onedot}
\def\etal{et~al\onedot} \def\iid{i.i.d\onedot}
\def\Fig{Fig\onedot} \def\Eqn{Eqn\onedot} \def\Sec{Sec\onedot} \def\Alg{Alg\onedot}
\makeatother

\newcommand{\xdownarrow}[1]{%
  {\left\downarrow\vbox to #1{}\right.\kern-\nulldelimiterspace}
}

\newcommand{\xuparrow}[1]{%
  {\left\uparrow\vbox to #1{}\right.\kern-\nulldelimiterspace}
}

\renewcommand\UrlFont{\color{blue}\rmfamily}

\newcommand*\rot{\rotatebox{90}}
\newcommand{\boldparagraph}[1]{\vspace{0.2cm}\noindent{\bf #1:} }
\newcommand{\boldquestion}[1]{\vspace{0.2cm}\noindent{\bf #1} }

\newcommand{\ftm}[1]{ \noindent {\color{magenta} {\bf Fatma:} {#1}} }
\newcommand{\ga}[1]{ \noindent {\color{blue} {\bf Gorkay:} {#1}} }

\newcommand{\supptitle}[1]{
   \newpage
   \null
  \vskip .375in
   \begin{center}
      {\Large \bf #1 \par}
      \vspace*{24pt}
      {
      \large
      \lineskip .5em
      \par
      }
   \end{center}
   }

\lstset{
  language=Python,
  commentstyle={\color{teal}\itshape},
  keywordstyle={\color{Green}\bfseries},
  stringstyle=\color{red},
  emph={predict_trajectory},
  emphstyle=\color{blue},
  basicstyle={\fontsize{7pt}{6pt}\selectfont\ttfamily}
}
\maketitle
\ificcvfinal\thispagestyle{empty}\fi

\begin{abstract}
Forecasting future trajectories of agents in complex traffic scenes requires reliable and efficient predictions for all agents in the scene. However, existing methods for trajectory prediction are either inefficient or sacrifice accuracy. To address this challenge, we propose ADAPT, a novel approach for jointly predicting the trajectories of all agents in the scene with dynamic weight learning. Our approach outperforms state-of-the-art methods in both single-agent and multi-agent settings on the Argoverse and Interaction datasets, with a fraction of their computational overhead. 
We attribute the improvement in our performance: first, to the adaptive head augmenting the model capacity without increasing the model size; second, to our design choices in the endpoint-conditioned prediction, reinforced by gradient stopping. 
Our analyses show that ADAPT can focus on each agent with adaptive prediction, allowing for accurate predictions efficiently. \url{https://KUIS-AI.github.io/adapt}
\end{abstract}
\section{Introduction}

A self-driving agent needs to be able to anticipate the future behavior of other agents around it to plan its trajectory. This problem, known as trajectory forecasting, is an important requirement for safe navigation.
There are multiple challenges to solving this problem. First of all, traffic scenes are highly dynamic. The behavior of an agent depends not only on the scene properties, such as configurations of lanes but also on other agents, such as yielding to another vehicle that has priority. Second, multiple futures need to be predicted due to the inherent uncertainty in future predictions. While these two challenges are studied in the literature, one challenge remains mostly unresolved: The future is shaped according to \textit{all} agents in the scene acting together. Therefore, trajectories of all agents need to be predicted as opposed to the current practice of predicting only the trajectory of a selected agent~\cite{Chang2019CVPR, Caesar2020CVPR}.

\begin{figure}[t!]
    \centering
        \begin{subfigure}{0.25\textwidth}
           \includegraphics[width=0.95\linewidth, trim={0.4cm 0cm 
           1.5cm 1.5cm}, clip]{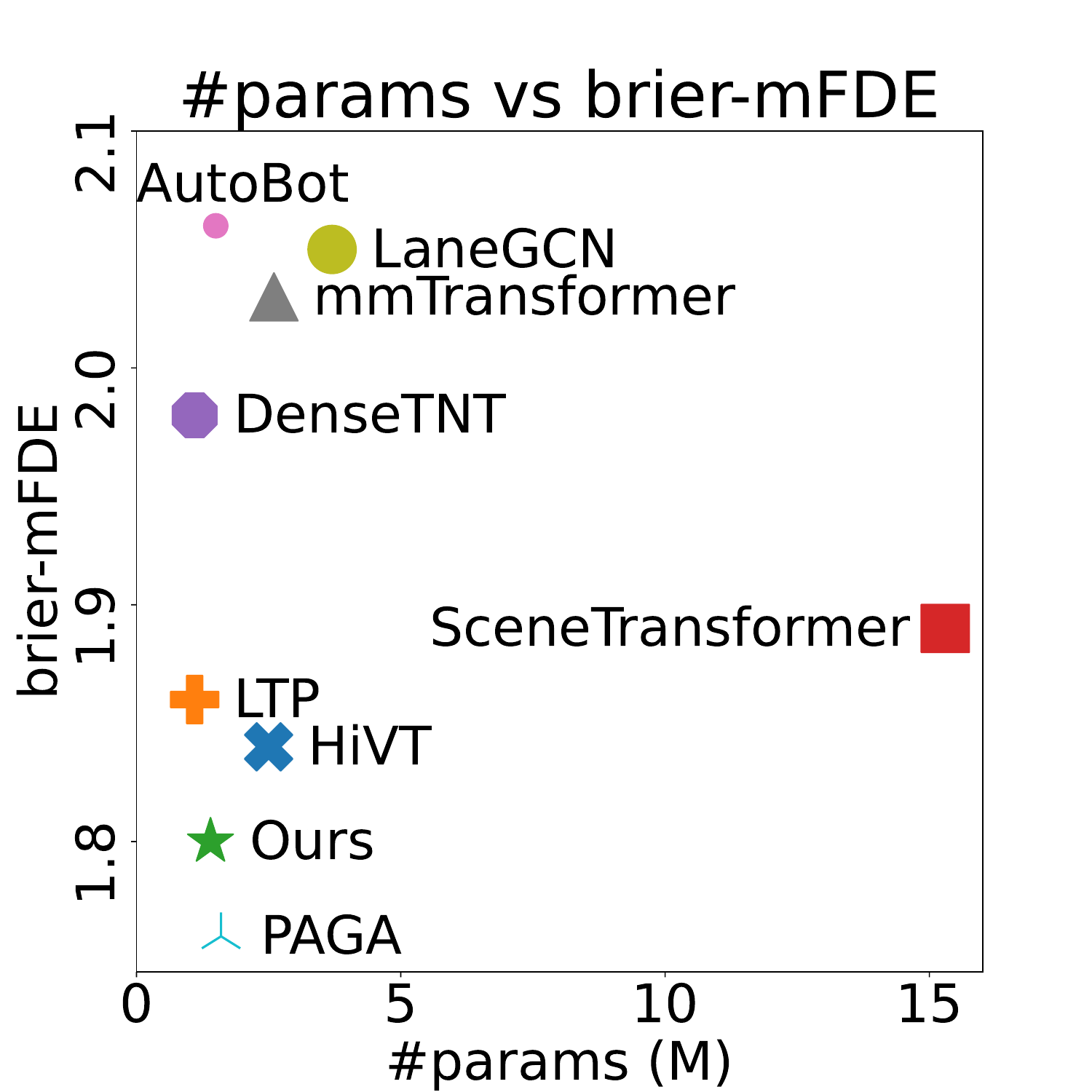}
           \caption{}
           \label{fig:param_v_brier} 
        \end{subfigure}%
        \begin{subfigure}{0.25\textwidth}
            \includegraphics[width=0.95\linewidth, trim={0.4cm 0.4cm 1.8cm 1.5cm}, clip]{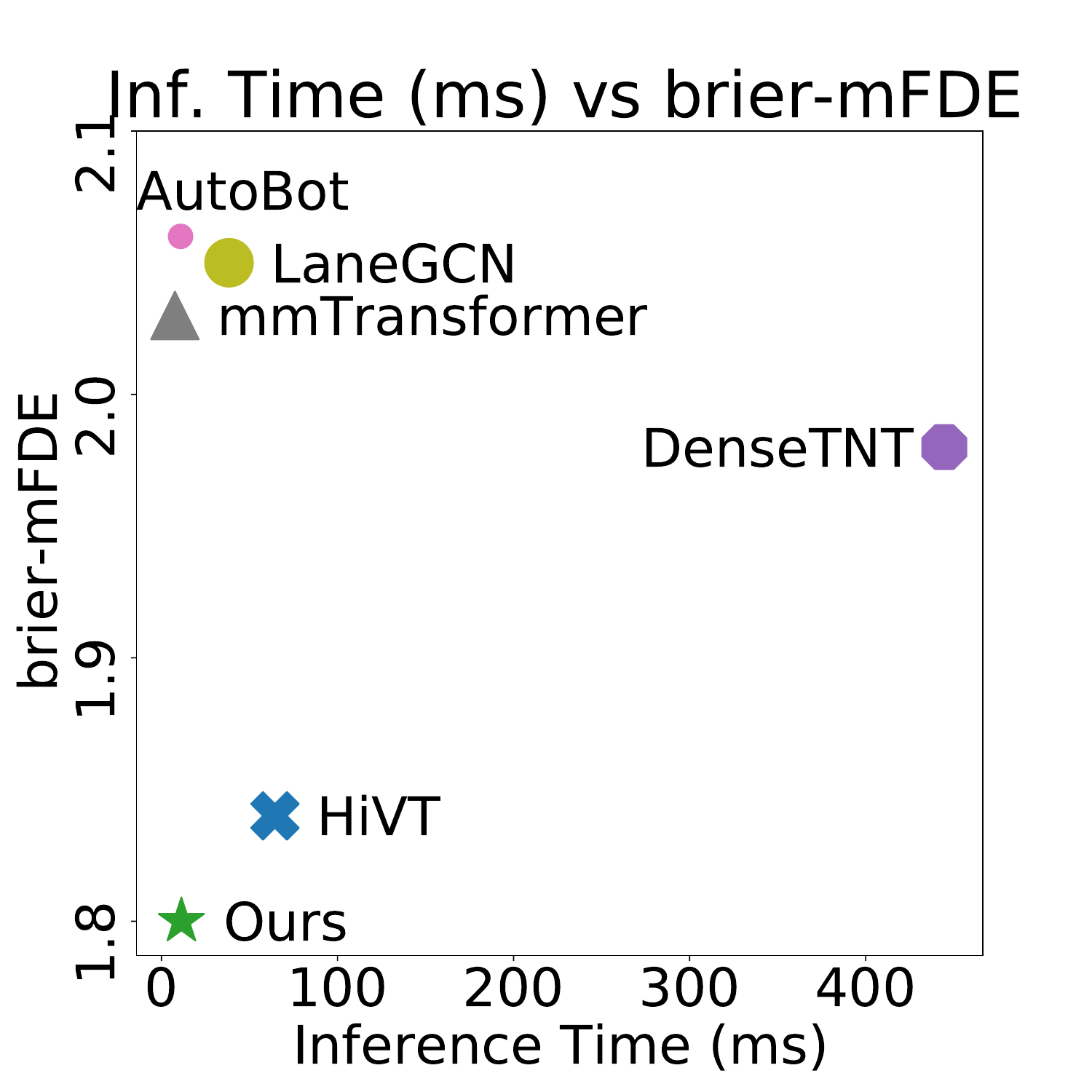}
            \caption{}
            \label{fig:time_v_brier} 
        \end{subfigure}
    \caption{\textbf{Accuracy vs. Efficiency.} We plot the accuracy in terms of error (brier-mFDE) vs. the number of parameters~(\subref{fig:param_v_brier}) and inference time~(\subref{fig:time_v_brier}) on the test set of Argoverse dataset~\cite{Chang2019CVPR}. Our method achieves one of the lowest reported errors with a small number of parameters, leading to highly efficient inference time compared to methods AutoBot~\cite{Girgis2022ICLR}, LaneGCN~\cite{Liang2020ECCV}, mmTransformer~\cite{Liu2021CVPR}, DenseTNT~\cite{Gu2021ICCV}, SceneTransformer~\cite{Ngiam2022ICLR}, LTP~\cite{Wang2022CVPR}, PAGA~\cite{Da2022ICRA}, and HiVT~\cite{Zhou2022CVPR}.}
\label{fig:acc_vs_eff}
\vspace{-0.4cm}
\end{figure}

\begin{figure}[t!]
    \centering
        \begin{subfigure}[t]{0.2\textwidth}
           \includegraphics[width=0.95\linewidth, trim={0 5cm 0 0}, clip]{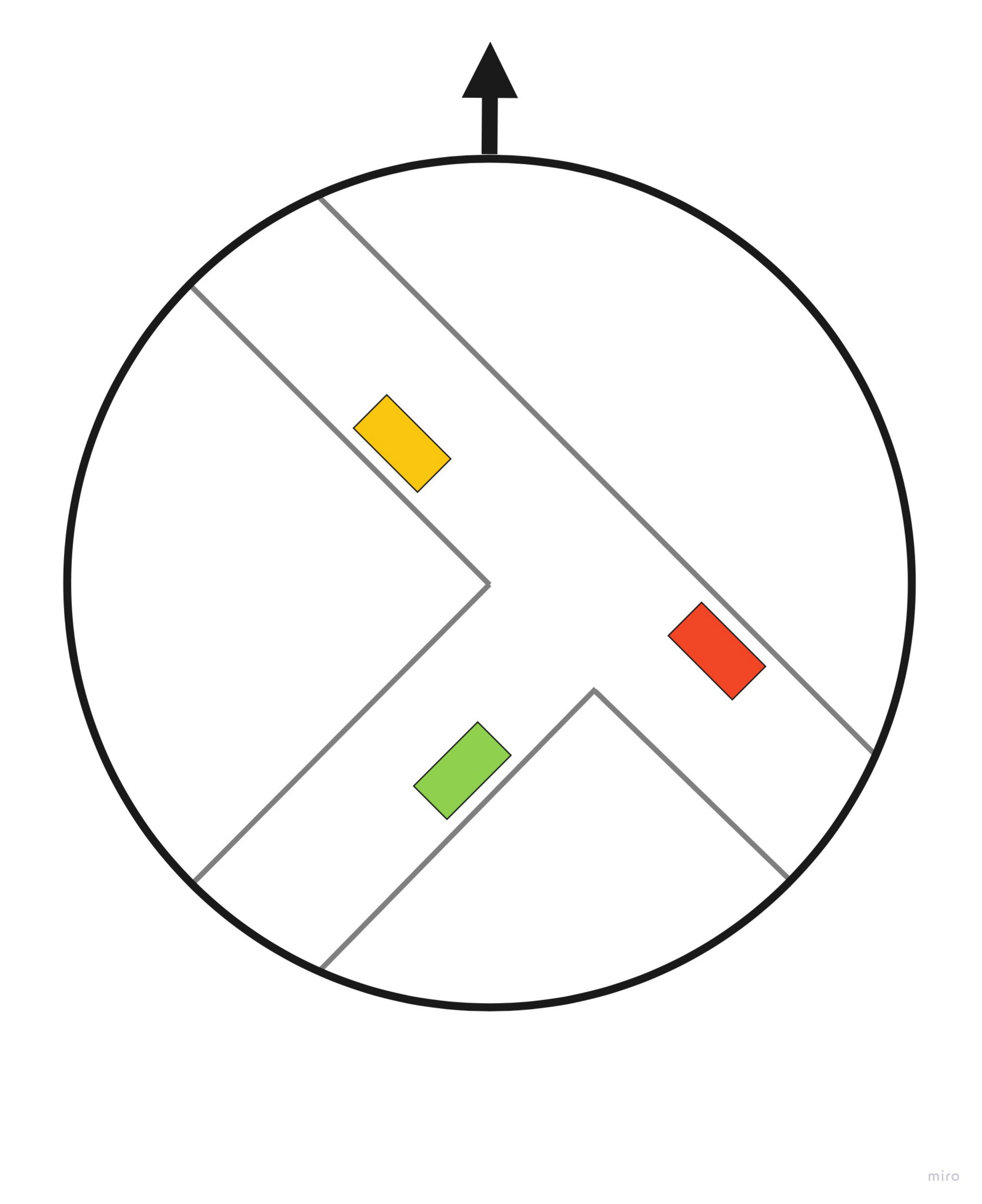}
           \caption{Scene-Centric}
           \label{fig:scene_centric} 
        \end{subfigure}%
        \begin{subfigure}[t]{0.3\textwidth}
            \includegraphics[width=0.95\linewidth, trim={0 2cm 0 3cm}, clip]{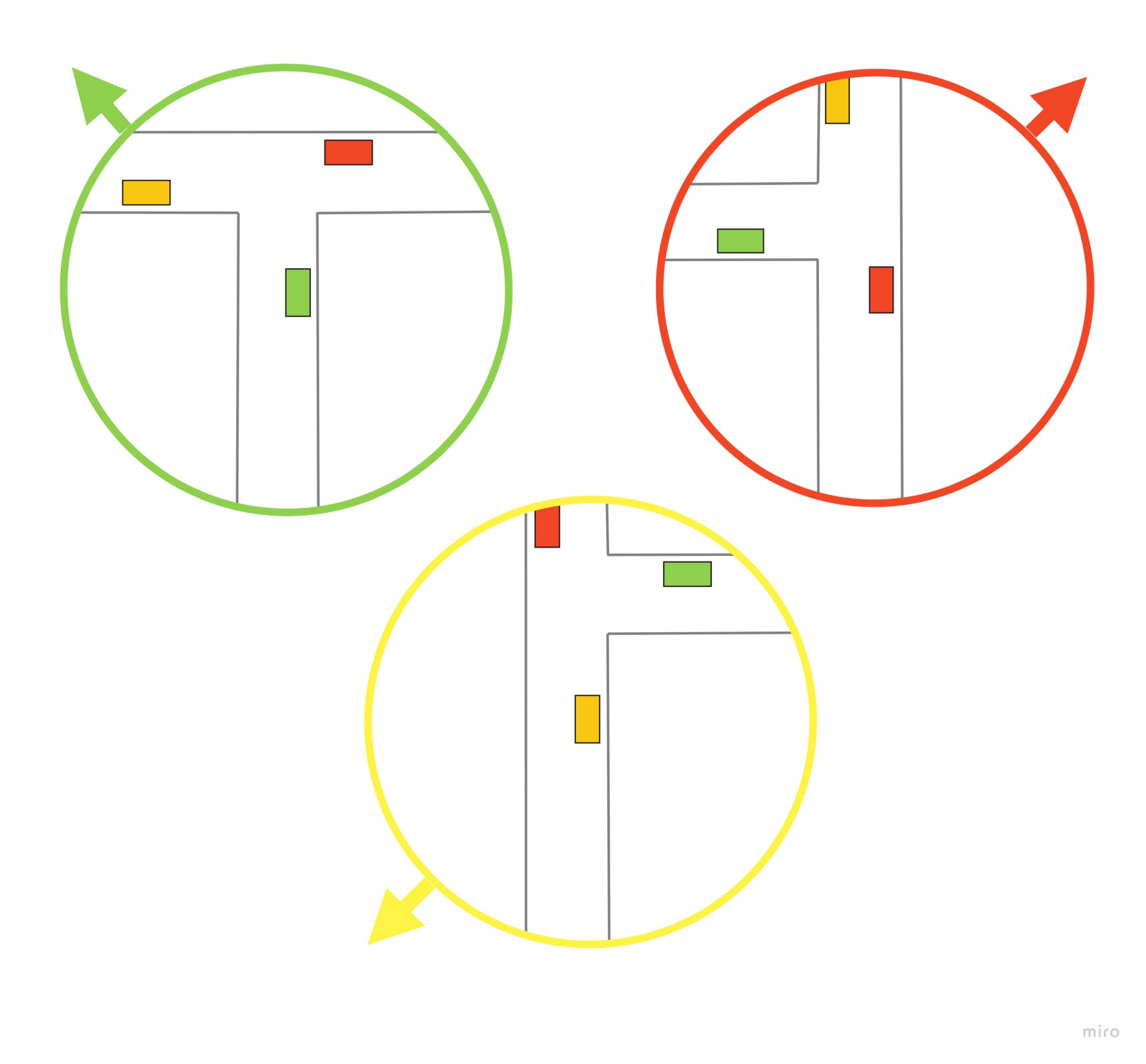}
            \caption{Agent-Centric}
            \label{fig:agent_centric} 
        \end{subfigure}
\caption{\textbf{Scene-Centric vs. Agent-Centric Representation.} In scene-centric representation~(\subref{fig:scene_centric}), all elements are encoded according to the same reference point. In agent-centric representation~(\subref{fig:agent_centric}), each agent is encoded in its own reference frame, leading to a complexity linear in the number of agents in multi-agent prediction.}
\label{fig:scene_vs_agent}
\vspace{-0.4cm}
\end{figure}

The progress in trajectory forecasting has focused mainly on scene representations for predicting the trajectory of a single agent. Typically, the existing methods~\cite{Liang2020ECCV, Ye2021CVPR, Gilles2022ICLR} follow an agent-centric reference frame where the scene is centered around the agent of interest, and everything else is positioned relative to it. This way, the prediction network is provided with the same initial state regardless of the agent's location or orientation, providing \textit{pose-invariance}. 
In other words, the scene is observed from the viewpoint of the agent of interest. In multi-agent setting, each agent has a different view of the world, and one cannot be prioritized over another as in the case of the agent-centric approach. A straightforward extension of an agent-centric approach to multi-agent is iterating the process for each agent in its own reference frame~(\figref{fig:scene_vs_agent}). This is achieved by transforming the scene according to each agent to obtain pose-invariant features. However, this solution scales linearly with the number of agents and causes a variable inference time that cannot be afforded in the real-time setting of driving. As a solution, SceneTransformer~\cite{Ngiam2022ICLR} introduces a global frame that is shared across agents. In their scene-centric approach, all agents are positioned with respect to the same reference point but at the cost of pose-invariance.

Ideally, the pose-invariance is a desirable property but for multi-agent prediction, a scene-centric approach can be preferred in real world due to efficiency concerns~\cite{Ngiam2022ICLR}. Then the question is how do we avoid the problems of a scene-centric approach without sacrificing efficiency? In this paper, we propose a solution to adapt to the situation of each agent with dynamic weight learning~\cite{Jia2016NeurIPS, Tian2020ECCV, Sun2021CVPR}.
Dynamic networks can adjust the model structure based on input by adapting network weights according to the changes in the input states~\cite{Han2021PAMI}. %
Therefore, they are well-suited for the multi-agent prediction task where each agent has a different initial state. Additionally, dynamic networks are capable of expanding the parameter space without increasing computation cost, therefore meeting the real-time requirements of our task.
We learn the weights of the network that predicts the endpoints so that they can change and adapt to each agent's reference frame. %
With dynamic weights, we can efficiently adapt the prediction head to each agent in a scene-centric approach without iterating over agents.

Our method is not only the first to achieve multi-agent prediction accurately and efficiently in a scene-centric approach but also one of the smallest and fastest among all trajectory prediction models including single-agent prediction. 
Using a goal-conditioning approach, we can easily switch between single and multi-agent prediction settings. To further enhance the performance of our model, we employ gradient stopping to stabilize the training of trajectory and endpoint prediction. This technique enables us to achieve good performance by fully leveraging the capacity of a small decoder with simple MLP layers rather than a complex one.

We show that our method outperforms the state-of-the-art methods with a fraction of their parameters in both single-agent setting of the Argoverse~\cite{Chang2019CVPR} and multi-agent setting of the Interaction~\cite{Wei2019ARXIV}. On Interaction, specifically designed for evaluating multi-agent predictions, our method achieves a $1\%$ miss rate in comparison to $5\%$ which was the lowest achieved so far~\cite{Gilles2022ICLR}. Our contributions can be summarized as follows:
\begin{itemize}[topsep=0.5pt]
    \setlength\itemsep{0em}
    \item We propose a novel approach for predicting the trajectories of all agents in the scene. Our adaptive head can predict accurate trajectories by dynamically adapting to various initial states of multiple agents.
    \item We achieve state-of-the-art results efficiently with one of the smallest and fastest models including the ones in single-agent setting. We validate our design choices in endpoint prediction and trajectory prediction with gradient stopping for stabilized training.
    \item We have created a unified prediction process that can be used for both single and multi-agent settings with the same backbone by utilizing endpoint conditioning. Our method allows for easy switching between scene-centric and agent-centric reference frames, achieving state-of-the-art in both settings.
\end{itemize}

\section{Related Work}

\subsection{Single-Agent Prediction}
\boldparagraph{Scene Representation}
In dynamic traffic scenes, representing the scene elements and modeling interactions between them play a crucial role in performance. In single-agent prediction, previous works focus on the representation of the scene and interaction modeling from the viewpoint of the agent of interest. Earlier works~\cite{Cui2019ICRA, Casas2018CoRL, Hong2019CVPR, Biktairov2020NeurIPS, Konev2022ARXIV, Luo2018CVPR} create a rasterized image to represent both the context and the interactions. The previous locations of agents are typically encoded with sequential models such as RNNs~\cite{Mercat2020ICRA, Khandelwal2020ARXIV, Alahi2016CVPR, Gupta2018CVPR, Park2020ECCV, Salzmann2020ECCV}. Combining the two, the following works explore more specialized scene representations~\cite{Tang2019NeurIPS, Chai2019CoRL, Djuric2020WACV, Gilles2021ITSC, Rhinehart2019ICCV, Buhet2020CoRL, Park2020ECCV, Mo2020ARXIV}. In contrast to rasterized representation, Graph Neural Networks~(GNNs) enable a more explicit way of modeling interactions such as with a lane graph~\cite{Liang2020ECCV} or a vectorized representation of the scene~\cite{Gao2020CVPR}. Based on their success in capturing hierarchical representations, recent works continue adapting GNNs for interaction modeling~\cite{Zhao2020CoRL, Gu2021ICCV, Zeng2021IROS, Aydemir2022ARXIV}. Towards the same purpose, more recent works~\cite{Zhou2022CVPR, Liu2021CVPR, Song2021CoRL, Ngiam2022ICLR, Gilles2022ICLR}  use transformers with multi-head attention~\cite{Vaswani2017NeurIPS}, whose success has been proven repeatedly~\cite{Dosovitskiy2020ICLR, Caron2021ICCV, Devlin2018ARXIV, Carion2020ECCV}. We also use a scene representation based on multi-head attention.

\boldparagraph{Attention-Based Encoding}
Transformers are widely used for interaction modeling due to their ability to capture the interaction between different scene elements. 
VectorNet~\cite{Gao2020CVPR} uses the same attention weights for different types of elements, \ie agents and lanes. LaneGCN~\cite{Liang2020ECCV} categorizes interactions and uses a different type of attention for each, leading to a more specialized representation. Due to its success, the following works~\cite{Liu2021CVPR, Wang2022CVPR, Zhou2022CVPR} continue modeling different types of interactions. Recently, factorized attention~\cite{Ngiam2022ICLR, Girgis2022ICLR} has been proposed to model temporal relations between scene elements efficiently. Instead of performing attention on the entire set of agents and time steps, factorized attention separately processes each axis, \ie agent, time, and road graph elements. 
A similar factorization over time and agents is explored in Autobot~\cite{Girgis2022ICLR} with a smaller and more efficient architecture. 
We also use different types of attention to model different types of interactions but enable updated information flow between different scene elements with iterative updates.

\begin{figure*}[t!]
    \centering
    \includegraphics[width=.91\linewidth, trim={0cm 3cm 0cm 0.5cm}, clip]{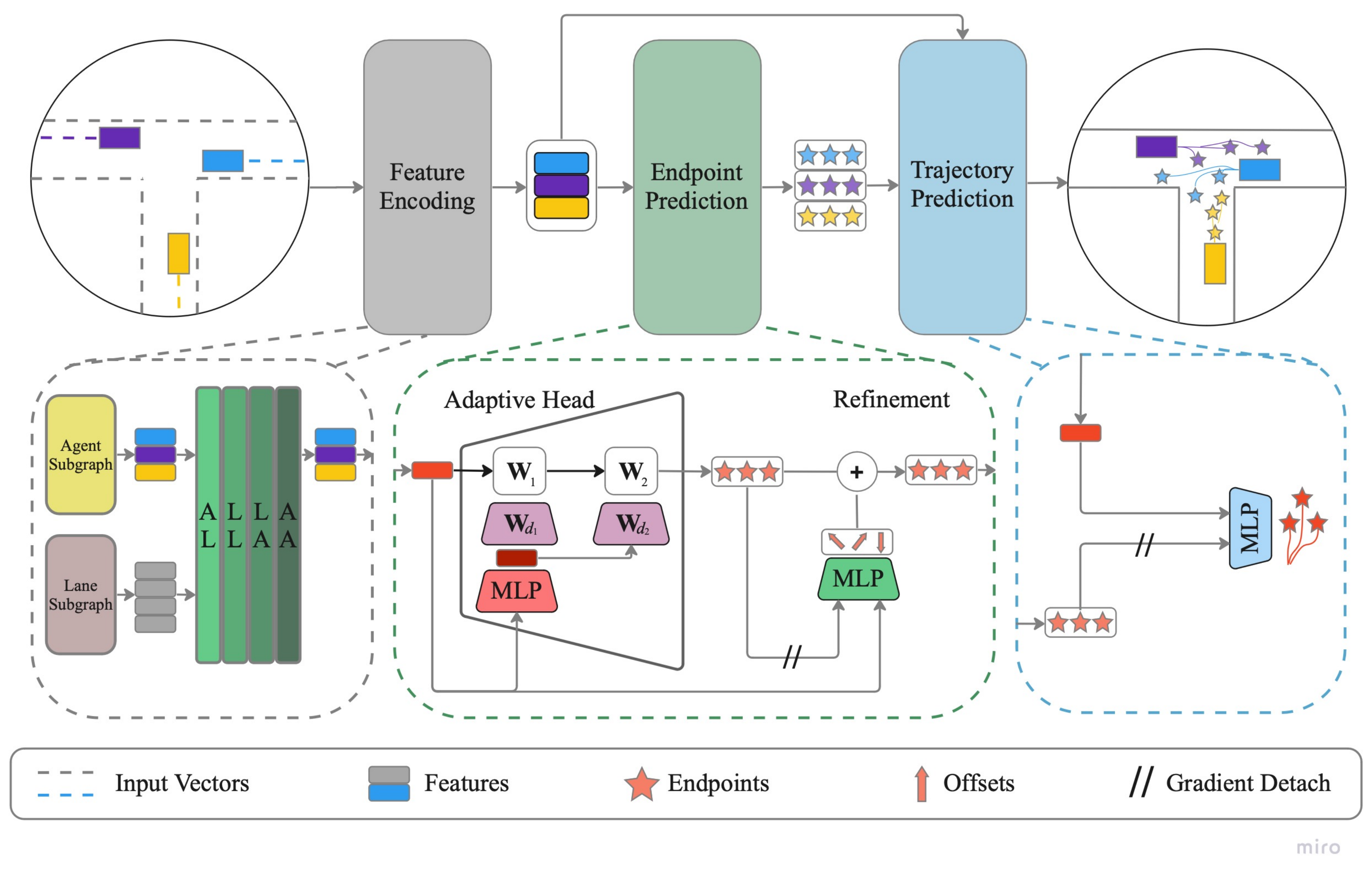}
    \caption{\textbf{Overview.} Our scene encoding approach involves separate polyline encoders that interact in feature encoding (\textbf{left}). To predict endpoint proposals, we utilize the endpoint head, which employs the adaptive head with dynamic weights for multi-agent prediction, without the need to transform the scene for each agent. Conversely, we use static head (simple MLP) for single-agent prediction. Then we perform endpoint refinement to improve accuracy (\textbf{middle}). Finally, we interpolate the full trajectory for each agent using the refined endpoints (\textbf{right}). By utilizing gradient detaching for endpoint and trajectory prediction modules, we achieve better performance with a small and fast architecture.}
    \label{fig:pipeline} 
    \vspace{-0.2cm}
\end{figure*}

\subsection{Multi-Agent Prediction}
\boldparagraph{Multi-Agent Prediction} 
Due to the evaluation setting on publicly available datasets~\cite{Chang2019CVPR, Caesar2020CVPR}, most existing works focus on predicting the trajectory of a single agent. While it led to great progress in scene representations and modeling of dynamic interactions, in real-life scenarios, the agent needs to account for the future trajectories of other agents as well. Multi-agent prediction has been recently addressed by SceneTransformer~\cite{Ngiam2022ICLR} with a unified architecture to jointly predict consistent trajectories for all agents. While this method can perform inference for all agents in a single forward pass, HiVT~\cite{Zhou2022CVPR} iterates over agents in an agent-centric representation, leading to the aforementioned inefficiency issues. LTP~\cite{Wang2022CVPR} follows a different approach with a lane-oriented representation to predict the most likely lane for each agent in a single pass. However, lane classification may not be as precise as regression. Our method can regress the trajectories of each agent efficiently in a single pass.

\boldparagraph{Reference Frame} 
The existing works represent the scene either from the viewpoint of an agent or from a fixed reference point as illustrated in \figref{fig:scene_vs_agent}. In the agent-centric representation~\cite{Zhou2022CVPR, Girgis2022ICLR, Ye2021CVPR, Gu2021ICCV, Zhao2020CoRL, Liang2020ECCV}, the scene is transformed so that the agent of interest is positioned at the origin of the scene. In contrast, all elements are positioned with respect to the same reference point in a scene-centric representation~\cite{Ngiam2022ICLR}. This shared context representation is especially helpful for multi-agent prediction~\cite{Wei2019ARXIV, Sun2020CVPR} while the agent-centric works better for single-agent prediction~\cite{Chang2019CVPR, Caesar2020CVPR} due to the simplification of the problem. It allows focusing on a single agent without worrying about other agents except for their relation to the agent of interest. Multi-agent prediction can be performed with sequential agent-centric predictions, \ie one agent at a time~\cite{Girgis2022ICLR, Wang2022CVPR, Gu2021ICCV, Zhou2022CVPR, Gilles2022ICLR}. However, this straightforward extension scales linearly with the number of agents in the scene and raises efficiency concerns. We use a scene-centric representation for multi-agent predictions but adapt to each agent with dynamic weights to benefit from agent-specific features as in agent-centric representation.

\section{Methodology}
Given the past trajectories of all agents on a High-Definition~(HD) map of the scene, our goal is to predict the future trajectories of agents in the scene. 
In a vectorized scene representation, we model different types of interactions between the agents and the map to obtain a representation for agents~(\secref{sec:feature_encoding}). Following goal-conditioned approaches~\cite{Zhao2020CoRL, Gu2021ICCV}, we first predict a possible set of endpoints. We then refine each endpoint by predicting an offset~(\secref{sec:endpoint_prediction}). Finally, we predict the full trajectories conditioned on endpoints~(\secref{sec:trajectory_prediction}). We stabilize training by separating endpoint and trajectory prediction with gradient detaching. Our pipeline is illustrated in~\figref{fig:pipeline}. Our model uses small MLPs in endpoint and trajectory prediction, keeping model complexity low.

\subsection{Feature Encoding}
\label{sec:feature_encoding}

\boldparagraph{Polyline Encoding} We represent the map and the agents using a vectorized representation in a structured way. The vectorized representation initially proposed in VectorNet~\cite{Gao2020CVPR} creates a connected graph for each scene element independently. Given past trajectories of agents, $\cA = \{\ba_i\}$ where $\ba_i \in \nR^{T \times 2}$ denotes the location of agent $i$ at previous $T$ time steps and HD map, $\cM = \{\bm_i\}$ where $\bm_i \in \nR^{l_i \times 2}$ denotes the lane $i$ with $l_i$ consecutive points constituting the lane. We encode each scene element, \ie a polyline, with a polyline subgraph. We use two separate subgraphs for the agents and lanes~(\figref{fig:pipeline}, left), resulting in a feature vector of length $d$, $\bff_i \in \mathbb{R}^{d}$ for each polyline.

\boldparagraph{Interaction Modelling} We model different types of interactions between the scene elements~(\figref{fig:pipeline}, left). Following LaneGCN~\cite{Liang2020ECCV}, we model four types of relations: agent-to-lane~(\textbf{AL}), lane-to-lane~(\textbf{LL}), lane-to-agent~(\textbf{LA}), and agent-to-agent~(\textbf{AA}).
Using attention, we update each feature $\bff_i$ extracted by the polyline subgraph. 

In contrast to previous work using simple attention operation~\cite{Liang2020ECCV}, and given the importance of multi-head attention and feed-forward networks 
in understanding the relations~\cite{Geva2020ARXIV},
we use a Multi-Head Attention Block~(MHAB) as proposed in \cite{Girgis2022ICLR}. 
Specifically, we update self-relations~(\textbf{AA, LL}) using a self-attention encoder followed by a feed-forward network~(FFN) and cross-relations~(\textbf{AL, LA}) using a cross-attention encoder followed by the FFN:
\begin{align}
    \text{MHA}(\bff_\bq,~\bff_{\bk\bv}) =&~ \text{softmax}\left(\frac{\bQ\,\bK^T}{\sqrt{dim_\bk}}\right)\bV \\
    \text{where}~~\bQ,\bK,\bV =&~ \bW^\bq\bff_\bq, \bW^\bk\bff_{\bk\bv}, \bW^\bv\bff_{\bk\bv} \nonumber
\end{align}
where each $\bW$ is a learned projection. Similar to the original block in~\cite{Vaswani2017NeurIPS}, our Multi-Head Attention Block is formally defined as follows:
\begin{align}
    \label{eq:mhab}
    \text{MHAB}(\bff_\bq,~\bff_{\bk\bv}) =&~ \text{norm}(\tilde{\bff} + \text{FFN}(\tilde{\bff})) \\
    \text{where}~~ \tilde{\bff} =&~ \text{norm}(\bff_\bq + \text{MHA}(\bff_\bq,~\bff_{\bk\bv})) \nonumber
\end{align}
where the norm is the Layer Normalization~\cite{Ba2016ARXIV}. Different than LaneGCN~\cite{Liang2020ECCV} applying different interaction types $L$ times sequentially one by one, we model each interaction in order and repeat the process $L$ times. This way, intermediate features can be updated at each iteration, and then the updated features are used to compute attention in the next iteration. Each scene element can be informed by different types of relations $L$ times. See Supplementary for the experiment comparing the two design choices.

\subsection{Endpoint Prediction}
\label{sec:endpoint_prediction}

For endpoint prediction, we either use a single MLP if an agent-centric reference frame is used which might be preferred due to its advantages in single-agent prediction %
or an adaptive head with dynamic weights if a scene-centric reference frame is used which might be preferred due to its efficiency in multi-agent prediction. %
Our model uses simple linear layers for endpoint prediction rather than sophisticated modules used in previous work~\cite{Ngiam2022ICLR, Girgis2022ICLR}.

\boldparagraph{Endpoint Prediction Head} We predict a possible set of endpoints for each agent based on the agent features from previous attention layers. We utilize two different types of heads to predict the future trajectory of a single agent in an agent-centric reference frame and future trajectories of multiple agents in a scene-centric frame. In single-agent setting, we predict the endpoints with a simple MLP, which we call a \textit{static head}. In multi-agent setting, we train an \textit{adaptive head} to dynamically learn the weights that predict the endpoints. Dynamic weight learning~\cite{Jia2016NeurIPS, Tian2020ECCV} enables the prediction head to adapt to the situation of each agent.
\begin{align}
    \label{eq:dynamic}
    \bW_1 &= \bW_{d_1} \tilde{\bff} \\
    \bW_2 &= \bW_{d_2} \tilde{\bff} \nonumber \\
    \bF_{d_1} &= \text{ReLU}(\text{norm}(\bW_1 \bff)) \nonumber \\
    \hat{\by}_{\text{pred}} &= \bW_2 \bF_{d_1} \nonumber
\end{align}



    


    

\begin{figure}
\begin{lstlisting}
def predict_trajectory(features):
    # features.shape = (N, D)

    features = features.expand(N, K, D)

    # endpoints.shape = (N, K, 2)
    endpoints = endpoint_head(features)
    
    # offsets.shape = (N, K, 2)
    offsets = MLP1(cat(features, endpoints.detach()))
    endpoints += offsets

    # traj.shape = (N, K, T-1, 2)
    traj = MLP2(cat(features, endpoints.detach()))
    traj = cat(traj, endpoints)

    # scores.shape = (N, K)
    scores = MLP3(cat(features, endpoints.detach()))
    scores = softmax(scores, dim=-1)
    
    return traj, scores
\end{lstlisting}
\vspace{-0.25cm}
\caption{\textbf{Pseudo-code for Trajectory Prediction.} Given features of $N$ agents, we first predict endpoint proposals using the endpoint head. Later, we apply a refinement on the endpoints by adding an offset. We then predict a trajectory for $T$ steps conditioned on each endpoint. We also predict a score associated with each trajectory.}
\label{fig:pseudo-code}
\vspace{-0.25cm}
\end{figure}

We visualize the adaptive head in \figref{fig:pipeline} and explain it mathematically in \eqref{eq:dynamic} where $\bW_{\bd_1}$ and $\bW_{\bd_2}$ are trainable parameters and norm is the layer normalization. We process the encoded agent features concatenated with meta info, $\bff$ with an MLP and obtain $\tilde{\bff}$. Meta info includes the direction and location information of the agent at prediction time. By providing the current state information to the prediction head as input, we allow the dynamic weights to adjust to the state of the agent while predicting the endpoints. %

\boldparagraph{Refinement} We further refine the endpoints by predicting an offset to the initial endpoint proposals from the prediction head. Given the endpoint proposals and the features of the agent, we predict the corresponding offset for each proposal with a simple MLP. We detach the gradients of endpoints before passing them as input to decouple the training of the endpoint prediction and refinement. Intuitively, offsets that are supposed to correct the endpoints can receive an independent gradient update from the endpoints. A similar approach is used to update queries in~\cite{Jia2022ICLR}.

\subsection{Trajectory Prediction}
\label{sec:trajectory_prediction}
\boldparagraph{Trajectory Interpolation} After obtaining the refined endpoint for each agent, we interpolate future coordinates between the initial point and the endpoint with an MLP. We detach the endpoints to ensure that weight updates for full trajectory prediction are separated from endpoint prediction. %
Similarly, we predict a probability for each trajectory using detached endpoints. %
We provide the pseudo-code for endpoint and trajectory prediction  in \figref{fig:pseudo-code}. We train static and adaptive heads as the endpoint head for the agent-centric and the scene-centric reference frames, respectively.

\boldparagraph{Training} For training, we predict $K$ trajectories and apply variety loss to capture multi-modal futures by back-propagating the loss only through the most accurate trajectory.
As we predict the full trajectories conditioned on the endpoints, the accuracy of endpoint prediction is essential for full trajectory prediction. 
Therefore, we apply a loss on endpoints to improve the endpoint prediction. The final term in our loss function is classification loss to guide the probabilities assigned to trajectories.
In summary, we train our model using the endpoint loss, the full trajectory loss, and the trajectory classification loss.
Please see Supplementary for the details of our loss functions.

\begin{figure*}[t!]
    \centering
    \begin{subfigure}[t]{0.6\textwidth}
       \includegraphics[width=0.98\linewidth, trim={3cm 1.1cm 2.4cm 1.3cm}, clip]{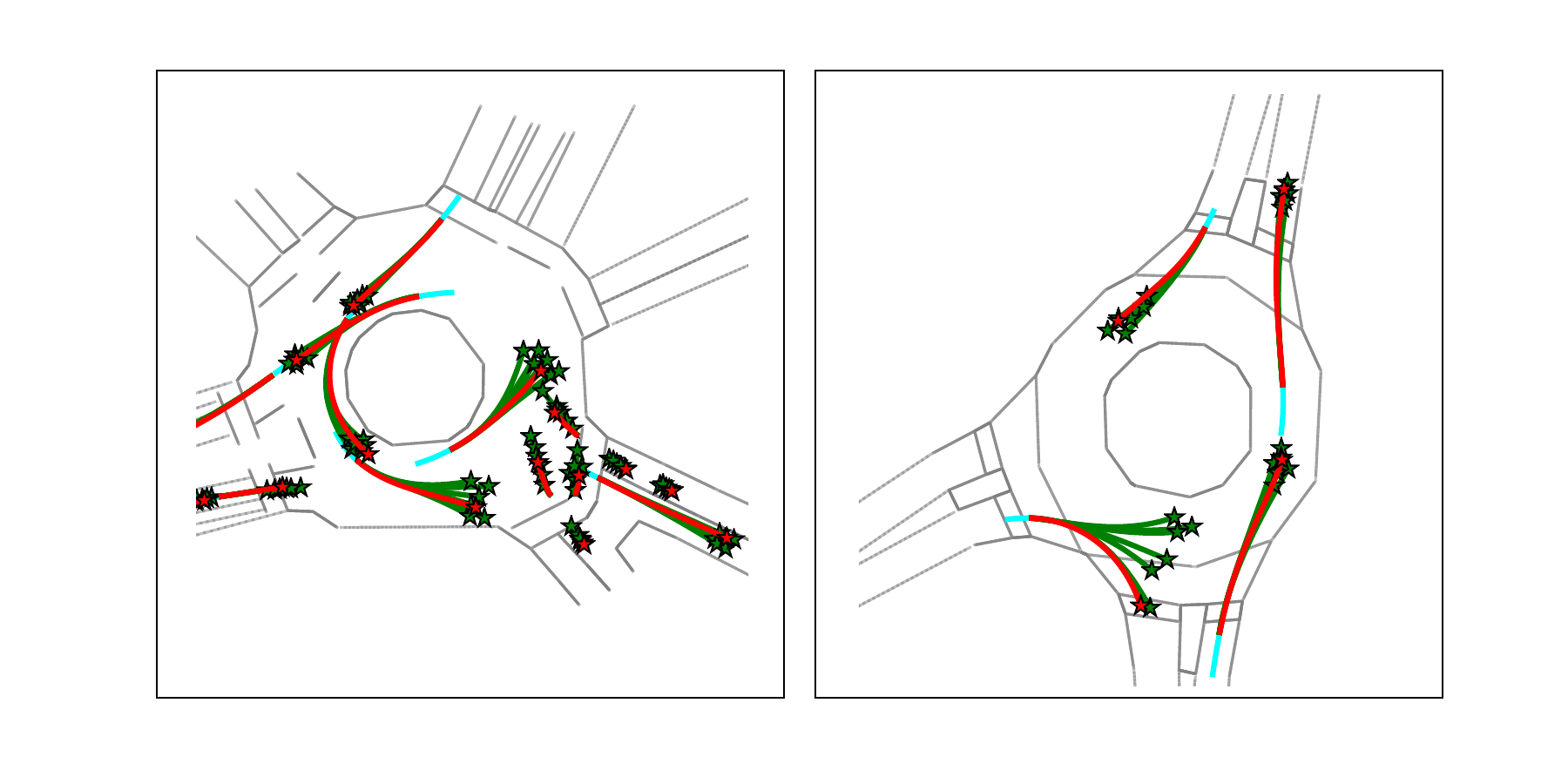}
       \caption{Multi-Agent}
       \label{fig:interaction_qual} 
    \end{subfigure}%
    \begin{subfigure}[t]{0.45\textwidth}
        \includegraphics[width=0.96\linewidth, trim={3.3cm 1.25cm 2.5cm 2cm}, clip]{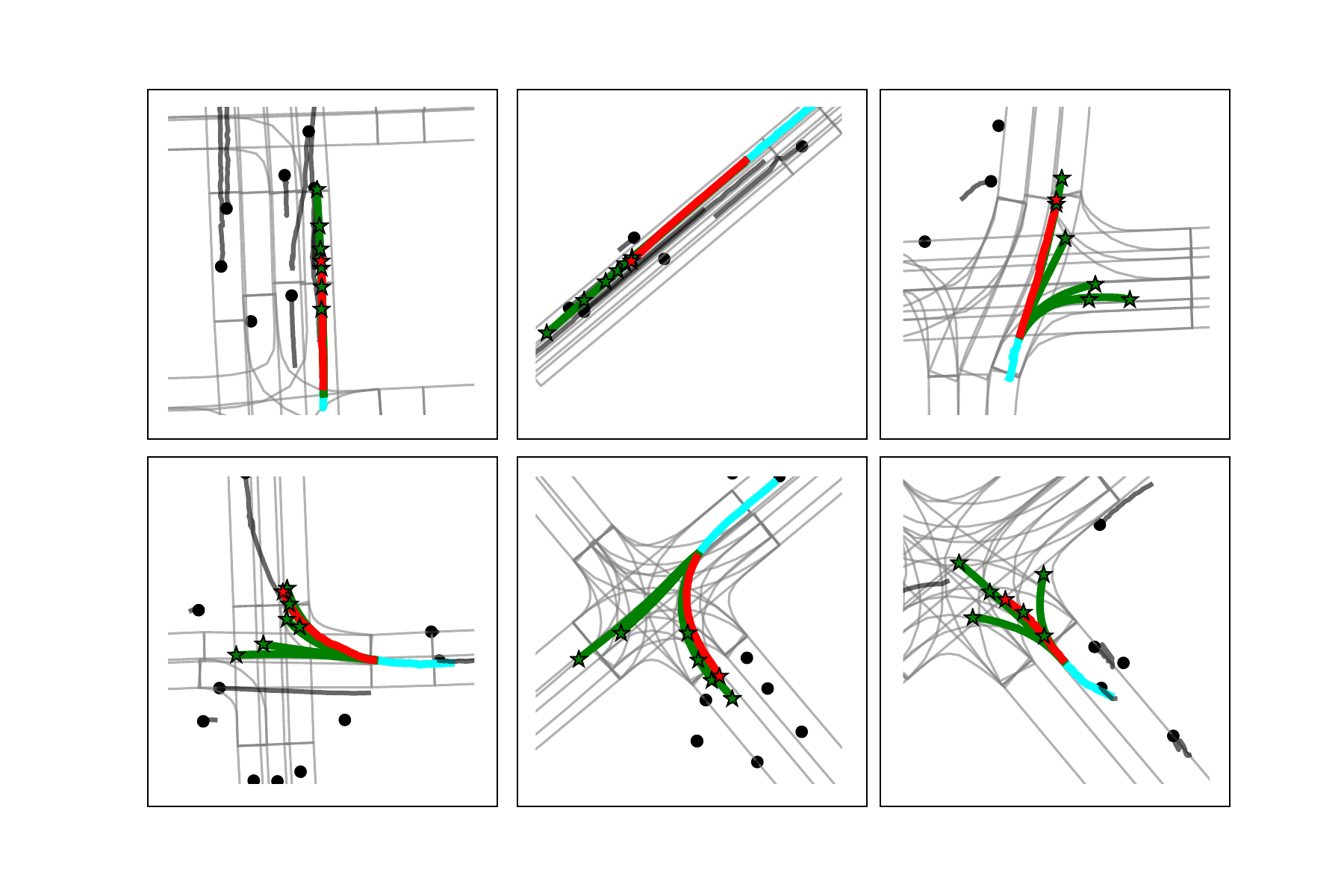}
        \caption{Single-Agent}
        \label{fig:argoverse_qual} 
    \end{subfigure}
\vspace{-.1cm}
\caption{\textbf{Qualitative Results.} We visualize multi-agent predictions on Interaction~(\subref{fig:interaction_qual}) and single-agent on Argoverse~(\subref{fig:argoverse_qual}). The predicted trajectories are shown in green, ground truth in red, past in cyan, and the trajectories of other agents in black.}
\vspace{-.1cm}
\label{fig:qualitative}
\end{figure*}
\section{Experiments}
\subsection{Experimental Setup}
\boldparagraph{Datasets} We evaluate our method in single-agent setting on Argoverse v1.1~\cite{Chang2019CVPR} and in multi-agent setting on Interaction~\cite{Wei2019ARXIV}. Argoverse, with 323,557 scenarios, is the commonly used benchmark for single-agent motion forecasting. Given the HD map and the history of agents for 2s, the goal is to predict the future locations of the agent of interest for the next 3s. Interaction contains 62,022 multi-agent scenarios with up to 40 agents per scenario. The goal is to predict the future for all agents in the scene. 

\boldparagraph{Metrics} We use standard metrics including minimum Average Displacement Error~($\text{mADE}_k$), minimum Final Displacement Error~($\text{mFDE}_k$), Miss Rate~($\text{MR}_k$), and brier minimum Final Displacement Error~(brier-$\text{mFDE}_k$). These metrics are calculated based on the trajectory with the closest endpoint to the ground truth over $k$ trajectory predictions. $\text{mADE}_k$ measures the average $\ell_2$ difference between the full prediction and the ground truth, $\text{mFDE}_k$ measures the difference between the predicted endpoint and the ground truth. $\text{MR}_k$ is the ratio of scenes where $\text{mFDE}_k$ is higher than 2 meters. The brier-$\text{mFDE}_k$ is calculated as $(1 - p)^2 + $ $\text{mFDE}_k$ where $p$ is the probability predicted for the trajectory. In multi-agent setting, each metric is computed per agent and then averaged over all agents.

\boldparagraph{Training Details}  We set the number of layers $L$ to 3 for both polyline subgraphs and interaction modeling. We train our models with a batch size of 64 for 36 epochs. We use Adam optimizer~\cite{Kingma2015ICLR} with an initial learning rate of $1 \times 10^{-4}$ and $2 \times 10^{-4}$ for Argoverse and Interaction experiments, respectively. We anneal the learning rate with a factor of $0.15$ at the $70^{th}$ and $90^{th}$ percentiles. We generate lane vectors for lanes that are closer than $50$m to any available agent. For data augmentation, we use random scaling in the range of $[0.75, 1.25]$  for Argoverse and random agent drop with the probability of $0.1$ for both Argoverse and Interaction experiments. We also use other agents on Argoverse as additional training data as done in previous work~\cite{Zhou2022CVPR}. Specifically,  we only consider agents that move by at least $6$m following previous work~\cite{Girgis2022ICLR, Ngiam2022ICLR}. In agent-centric reference frame, we translate and rotate the scene with respect to the agent of interest. In scene-centric reference frame, we orient the scene based on the mean location of all agents.

\subsection{Quantitative Results}
\begin{table}
  \centering
  \small
  \begin{tabular}{l | c c | c c | c | c }
    \toprule
     & \multicolumn{2}{c|}{$\text{mADE}_\text{k}$}  & \multicolumn{2}{c|}{$\text{mFDE}_\text{k}$} & \text{brier-} &  {\#Prm} \\
     \cline{2-3} \cline{4-5} & $\text{k=1}$ & \text{k=6} & $\text{k=1}$ & \text{k=6} & $\text{mFDE}_\text{6}$ & (M)\\
     
    \midrule
    
    AutoBot~\cite{Girgis2022ICLR} & - & 0.89 & - & 1.41 & - & 1.5 \\
    HO+GO~\cite{Gilles2022ICRA} & - & 0.92 & 3.68 & 1.29 & - & - \\
    LaneGCN~\cite{Liang2020ECCV} & 1.71 & 0.87 & 3.78 & 1.36 & 2.05 & 3.7 \\
    mmTr~\cite{Liu2021CVPR} & 1.77 & 0.84 & 4.00 & 1.34 & 2.03 & 2.6 \\
    D-TNT~\cite{Gu2021ICCV} & 1.68 & 0.88 & 3.63 & 1.28 & 1.98 & \textbf{1.1} \\
    THOMAS~\cite{Gilles2022ICLR} & 1.67 & 0.94 & 3.59 & 1.44 & 1.97 & - \\
    SceneTr~\cite{Ngiam2022ICLR} & 1.81 & 0.80 & 4.05 & 1.23 & 1.89 & 15.3 \\
    LTP~\cite{Wang2022CVPR} & 1.62 & 0.83 & 3.55 & 1.30 & 1.86 & \textbf{1.1} \\
    HiVT~\cite{Zhou2022CVPR} & 1.60 & \textbf{0.77} & 3.53 & \textbf{1.17} & 1.84 & 2.5 \\
    MP++*~\cite{Varadarajan2022ICRA} & 1.62 & 0.79 & 3.61 & 1.21 & 1.79 & 125 \\
    PAGA~\cite{Da2022ICRA} & \textbf{1.56} & 0.80 & \textbf{3.38} & 1.21 & \textbf{1.76} & 1.6 \\
    Ours~(Single) & \underline{1.59} & \underline{0.79} & \underline{3.50} & \textbf{1.17} & \underline{1.80} & \underline{1.4} \\
    \bottomrule
  \end{tabular}
  \caption{\textbf{Results on Argoverse (Test).} This table shows single-agent results on Argoverse. The number of parameters is reported or calculated using the official repositories. Models with ensembles are marked with ``*".}
  \label{tab:sota_argoverse_test}
  \vspace{-0.25cm}
\end{table}
\boldparagraph{Single-Agent Prediction on Argoverse}
We compare our method in single-agent setting using an agent-centric reference frame to the state-of-the-art on test~(\tabref{tab:sota_argoverse_test}) and validation~(\tabref{tab:sota_argoverse_val}) sets of Argoverse. 
We report results without ensembles, except for Multipath++~\cite{Varadarajan2022ICRA} (only the result of the ensemble is reported). Our method achieves comparable results to the top-performing methods on the test set in all metrics. In particular, we approach the the performance of the state-of-the-art PAGA~\cite{Da2022ICRA} in the official metric, \ie brier-$\text{mFDE}_6$ and reach the performance of HiVT~\cite{Zhou2022CVPR} in 
other metrics by using only $56\%$ of its parameters.
On the validation set, our method performs the best in terms of $\text{mFDE}_6$ and $\text{MR}_6$.
Impressively, ADAPT achieves these results with one of the smallest and fastest models~(\figref{fig:acc_vs_eff}). 

\begin{table}[b!]
  \centering
  \small
  \begin{tabular}{l | c c c | c}
    \toprule
     & $\text{mADE}_6$ & $\text{mFDE}_6$ & $\text{MR}_6$ & Inf. (ms) \\
    \midrule
    TPCN \cite{Ye2021CVPR} & 0.73 & 1.15 & 0.11 & - \\
    mmTrans \cite{Liu2021CVPR} & 0.71 & 1.15 & 0.11 & \textbf{7.66} \\
    LaneGCN \cite{Liang2020ECCV} & 0.71 & 1.08 & - & 38.37\\
    LTP \cite{Wang2022CVPR} & 0.78 & 1.07 & - & - \\
    DenseTNT \cite{Gu2021ICCV} & 0.73 & 1.05 & 0.10 & 444.66\\
    HiVT \cite{Zhou2022CVPR} & \textbf{0.66} & \underline{0.96} & \underline{0.09} & 64.45\\
    PAGA~\cite{Da2022ICRA} & 0.69 & 1.02 & - & -\\
    Ours (Single) & \underline{0.67} & \textbf{0.95} & \textbf{0.08} & \underline{11.31} \\
    \midrule
    Ours (Multi) & 0.65 & 0.97 & 0.08 & \underline{11.31} \\
    \bottomrule
  \end{tabular}
  \caption{\textbf{Results on Argoverse (Validation).} 
  This table shows results on the Argoverse validation set. The bottom row shows the multi-agent evaluation in scene-centric reference frame. The inference time is calculated using the official repositories in the same setting. }
  \label{tab:sota_argoverse_val}
\end{table}

In \tabref{tab:sota_argoverse_val}, we report the average runtime per scene on the validation set of Argoverse using a Tesla T4 GPU. To align the settings between different approaches and alleviate implementation differences in parallelism, we set the batch size to 1 and predict the future only for agent of interest per scene. Our method is the second fastest method, only behind mmTransformer~\cite{Liu2021CVPR} but with significantly better results than mmTransformer on both validation and test sets. Note that HiVT~\cite{Zhou2022CVPR} suffers significantly in terms of inference time due to agent-centric approach where the scene is normalized for each agent iteratively. Our approach can achieve similar results, and even slightly better, on the test set with only $18\%$ of HiVT's inference time. 
We provide a comparison of methods in terms of computational complexity in Supplementary to justify our design choices in feature encoding, contributing to our method's efficiency.
A full comparison in terms of brier-$\text{mFDE}_6$ vs. inference time is provided in \figref{fig:time_v_brier}. ADAPT achieves the best performance without sacrificing efficiency.

\boldparagraph{Multi-Agent Predictions on Argoverse} 
We extend the Argoverse setting from single-agent in agent-centric reference frame to multi-agent in scene-centric reference frame by modifying the reference point to be the same for all agents. In this case, we use ADAPT with adaptive head instead of static head. For evaluating multi-agent predictions, we only consider the agents that are visible at all timestamps. As shown at the bottom row of \tabref{tab:sota_argoverse_val}, ADAPT can predict the future trajectories of all agents in scene-centric reference frame with similar accuracy to single-agent case which has the advantage of agent-centric reference frame. Please note that the inference time remains the same from single-agent to multi-agent case since we predict all future trajectories in a single pass.

\begin{table}[t!]
  \centering
  \small
  \begin{tabular}{l | c c c | c}
    \toprule
     & $\text{mADE}_6$ & $\text{mFDE}_6$ & $\text{MR}_6$ & Inf. (ms) \\
    \midrule
    AutoBot (J.)~\cite{Girgis2022ICLR} &  \underline{0.21} & 0.64 & 0.06 & \underline{25.29} \\
    SceneTr~\cite{Ngiam2022ICLR} & 0.26 & 0.47 & \underline{0.05} & - \\
    THOMAS~\cite{Gilles2022ICLR} & 0.26 & \underline{0.46} & \underline{0.05} & - \\
    Ours (Multi) & \textbf{0.16} & \textbf{0.34} & \textbf{0.01} & \textbf{11.10} \\
    \bottomrule
  \end{tabular}
  \caption{\textbf{Results on Interaction (Validation).} This table shows multi-agent results on Interaction. The inference time is calculated using the official repositories. The results of SceneTransformer are based on a re-implementation by authors of THOMAS~\cite{Gilles2022ICLR}.}
  \label{tab:interaction_sota}
  \vspace{-0.45cm}
\end{table}

\boldparagraph{Multi-Agent Prediction on Interaction}
We compare our method in multi-agent setting using a scene-centric reference frame to other methods on the Interaction validation set in \tabref{tab:interaction_sota}. Our method significantly outperforms other methods with a large gap in all metrics. Impressively, it reaches $1\%$ miss rate, showing that our method can predict future trajectories accurately for all agents in the scene. %

\begin{table}[b]
  \centering
  \small
  \begin{tabular}{l | c c c }
    \toprule
    $\sigma$ & $\text{mADE}_6$ & $\text{mFDE}_6$ & $\text{MR}_6$ \\
    \midrule
    0 & 0.161 & 0.344 & 0.010 \\
    0.4 & 0.161 & 0.347 & 0.010 \\
    0.8 & 0.163 & 0.349 & 0.010 \\
    1.6 & 0.166 & 0.357 & 0.010 \\
    3.2 & 0.169 & 0.361 & 0.011 \\
    \bottomrule
  \end{tabular}
  \vspace{-0.2cm}
  \caption{\textbf{Robustness of Adaptive Head on Interaction (Validation).} This table shows the effect of adding noise to input coordinates on multi-agent performance.} 
  \label{tab:meta_info_robustness}
  \vspace{-0.2cm}
\end{table}
\boldparagraph{Robustness of Adaptive Head} To evaluate the effect of noisy input data, we conducted an experiment in multi-agent setting on Interaction. Specifically, we perturb input coordinates with noise $\mathcal{N}(0, \sigma)$ where $\sigma \in \{0.4, 0.8, 1.6, 3.2\}$, corresponding to an average of $\{0.32, 0.64, 1.28, 2.56\}$ meters deviation in 0.1 seconds, respectively. As shown in \tabref{tab:meta_info_robustness}, the performance is quite robust to increasing noise levels in input coordinates.

\begin{figure}[t]
  \centering
  \begin{subfigure}[t]{\linewidth}
    \centering
    \includegraphics[width=0.8\linewidth, trim={3cm 0.7cm 2.5cm 0.7cm}, clip]{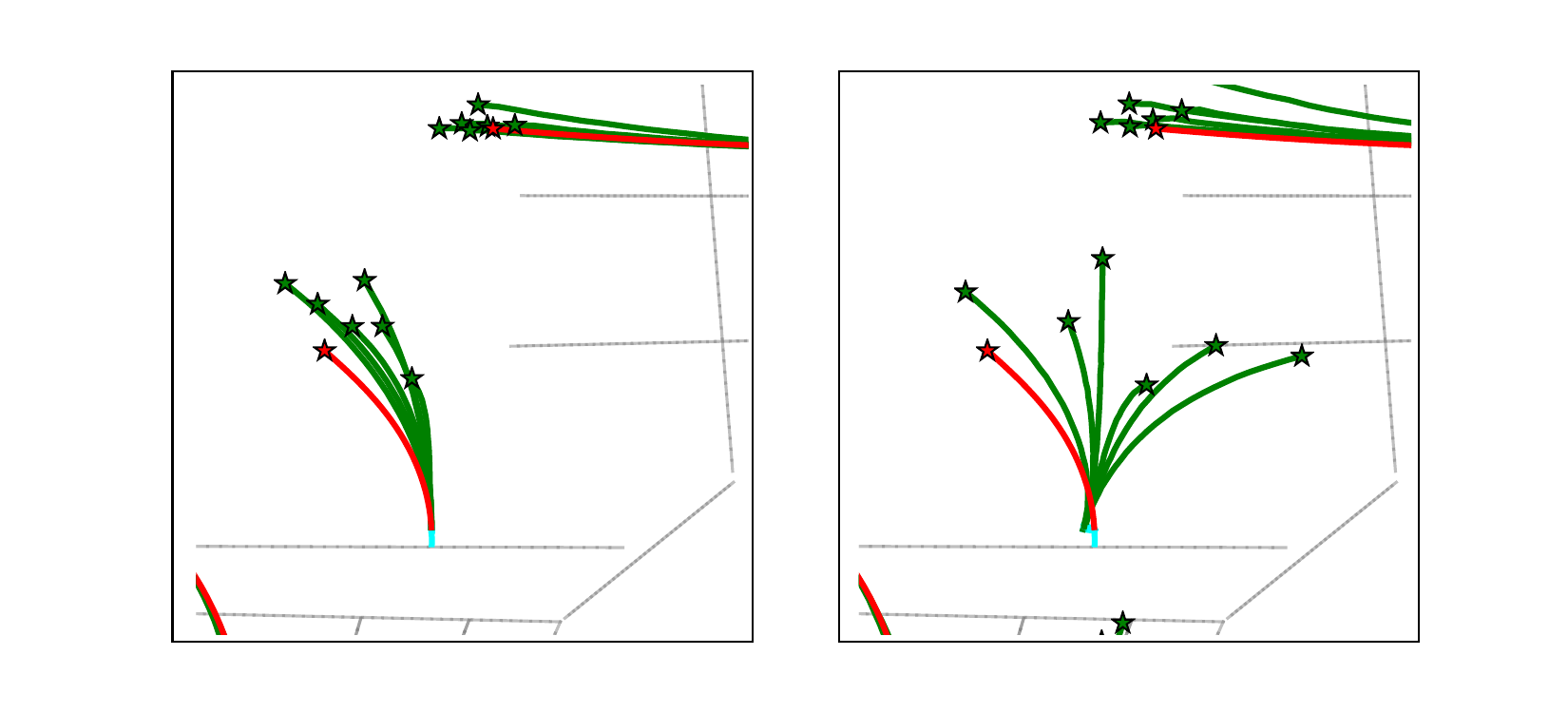}
  \end{subfigure}
  \hfill
  \begin{subfigure}[t]{\linewidth}
    \centering
    \includegraphics[width=0.8\linewidth, trim={3cm 0.7cm 2.5cm 0.7cm}, clip]{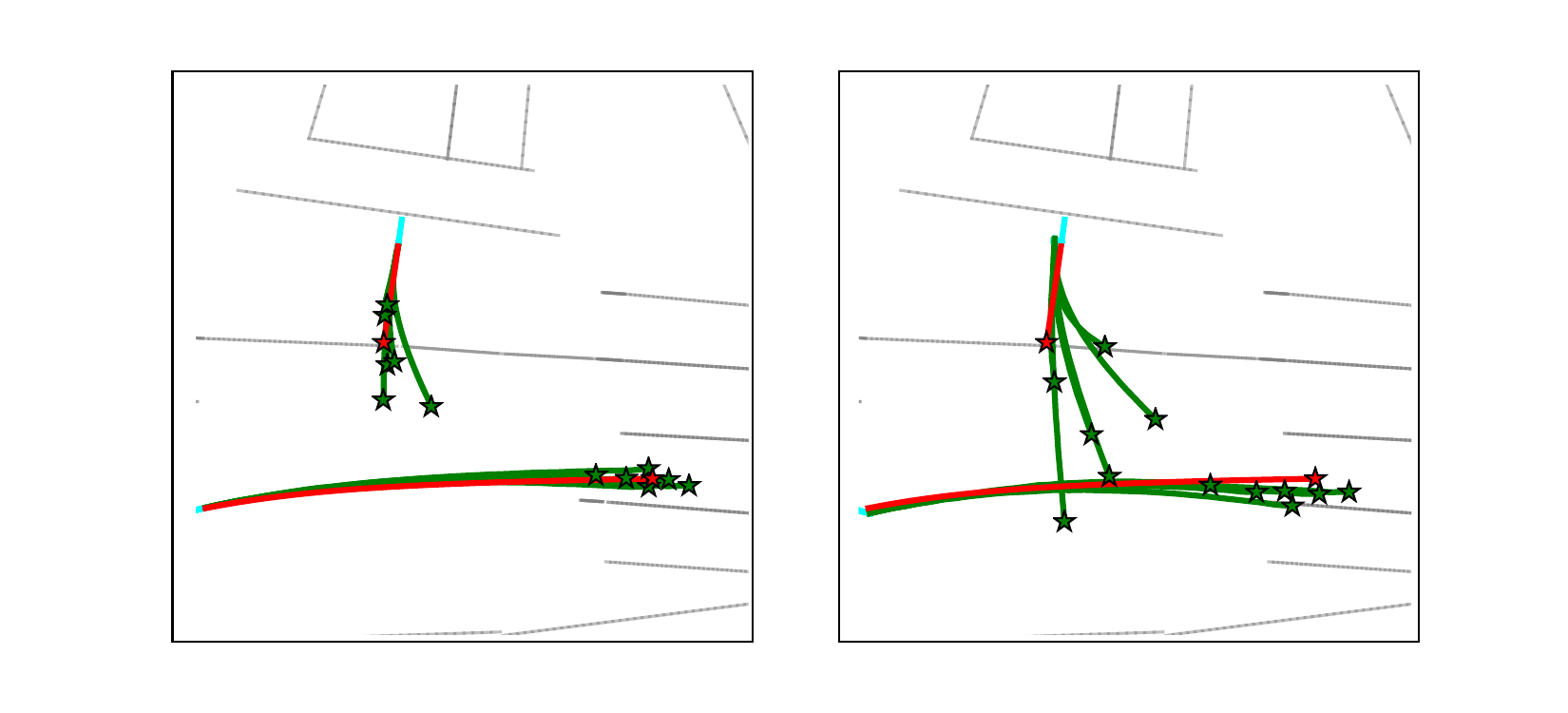}
  \end{subfigure}
  \caption{\textbf{Visualizating Effect of Adaptive Head.} We visualize the predictions of adaptive head (\textbf{left}) and static head (\textbf{right}). The predicted trajectories are shown in green, and ground truth in red.}
  \label{fig:adaptive_comparison}
  \vspace{-0.2cm}
\end{figure}

\subsection{Qualitative Analysis}
In \figref{fig:qualitative}, we visualize the predictions of our model in multi-agent setting on Interaction~(\subref{fig:interaction_qual}) and in single-agent setting on Argoverse~(\subref{fig:argoverse_qual}). Our model can predict accurate multi-modal trajectories for all agents in complex intersection scenarios on Interaction. In single-agent case, our model can vary predictions for the agent of interest in terms of intention and speed. Our model can successfully capture the interactions between the agents and predict futures by considering other agents, \eg on the top left in~\figref{fig:argoverse_qual}.

\boldparagraph{Visualizing Effect of Adaptive Head} To understand the importance of adaptive head, we compare the predictions of adaptive head (left) to the predictions of static head (right) in the same scenario in \figref{fig:adaptive_comparison}.
The adaptive head significantly improves the consistency and accuracy of predictions by allowing the model to adapt to the initial state of the agent, including its rotation, location, and speed.

\begin{figure}[t!]
    \centering
    \includegraphics[width=.9\linewidth, trim={3.5cm 2cm 10cm 2cm}, clip]{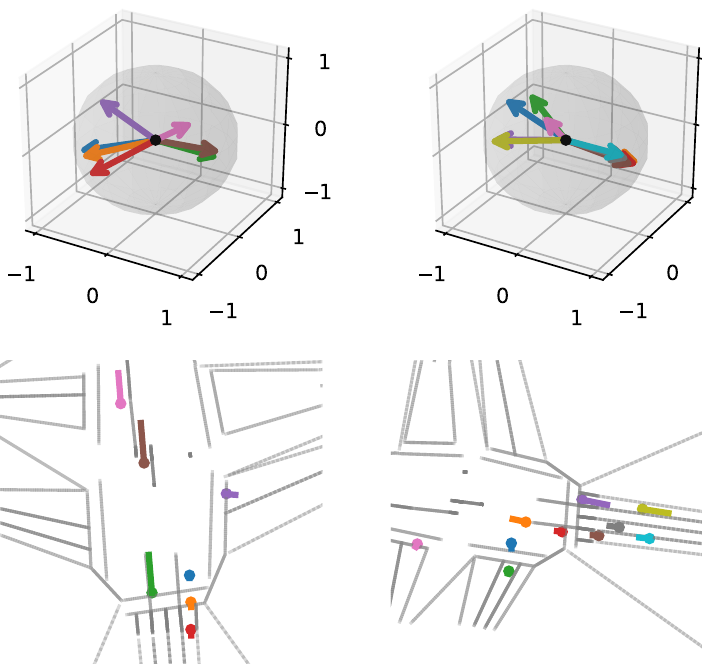}
    \caption{\textbf{Visualization of Dynamic Weights.} We project the dynamic weights for each agent into the 3D hypersphere (\textbf{top}) for two scenes on Interaction (\textbf{bottom}). We use the same color for the agents and their corresponding projections on the hypersphere.}
    \label{fig:w_hypersphere} 
    \vspace{-0.2cm}
\end{figure}

\boldparagraph{Understanding Dynamic Weights} To understand how the proposed adaptive prediction head changes according to each agent, we visualize dynamically learned weights for each agent by projecting them into the 3D hypersphere as shown in \figref{fig:w_hypersphere}. Specifically, we project the $\bW_2$ matrix~(\eqnref{eq:dynamic}) for each agent to a 3-dimensional vector with PCA and normalize it to unit length as shown on top of each scene. Despite differences in absolute position, the learned weights for agents moving in the same lane map to similar vectors on the hypersphere. For example, the brown and green agents on the left column map to almost identical vectors, although they are spatially far from each other. A separating factor is the orientation of agents, which is preserved in the mapping. For example, the green and purple agents on the right column map along the same direction, while the orange, red, brown, and cyan agents on the opposite lane map to the opposite direction.

\begin{table}[b]
   \begin{center}
      \small
      \centering
      \def\arraystretch{1.2}
      \begin{tabular}{l | c c c}
         \toprule
          & $\text{mADE}_6$ & $\text{mFDE}_6$ & $\text{MR}_6$ \\
         \midrule
         w/o Extended & 0.694 & 1.000 & 0.090 \\
         w/o Stop Gradient & 0.685 & 0.994 & 0.088 \\
         w/o Refinement & 0.683 & 0.990 & 0.088 \\
         w/o Augmentation & 0.675 & 0.974 & 0.089 \\
         ADAPT & \textbf{0.668} & \textbf{0.948} & \textbf{0.083} \\
         \bottomrule
      \end{tabular}
   \end{center}
   \vspace{-0.2cm}
   \caption{\textbf{Single-Agent Ablation Study on Argoverse (Validation).} This table shows the effect of removing each component on single-agent prediction on Argoverse.}
   \label{tab:ablation_single}
   \vspace{-0.25cm}
\end{table}

\subsection{Ablation Study}
\label{sec:ablation}
We conduct ablation studies on the validation split of the Argoverse for single-agent setting and the validation split of the Interaction dataset for multi-agent setting.

\begin{table}[t]
      \centering
      \small
      \def\arraystretch{1.2}
      \begin{tabular}{l | c c c}
         \toprule
         & $\text{mADE}_6$ & $\text{mFDE}_6$ & $\text{MR}_6$ \\
         \midrule
          w/o Adaptive Head & 0.244 & 0.425 & 0.017 \\
         ADAPT & \textbf{0.161} & \textbf{0.344} & \textbf{0.010} \\
         \bottomrule
      \end{tabular}
   \caption{\textbf{Multi-Agent Ablation Study on Interaction (Validation).}  This table shows the effect of using the adaptive head with dynamic weights for endpoint prediction on the performance of multi-agent prediction on Interaction.}
   \label{tab:ablation_multi}
   \vspace{-0.2cm}
\end{table}

\boldparagraph{Ablations on Single-Agent} In \tabref{tab:ablation_single}, we perform an ablation study on our architectural choices in single-agent setting of the Argoverse. First, using other agents in a scene as done in previous work~\cite{Zhou2022CVPR, Ngiam2022ICLR, Girgis2022ICLR} improves the performance in all metrics as it provides more samples for training. 
Second, gradient stopping in the trajectory predictor enhances the performance by providing independent updates for endpoint refinement and trajectory scoring. Third, refinement improves the accuracy of both the endpoint and the full trajectory by improving the accuracy of the initial endpoint as well.
Fourth, data augmentation increases the diversity of the training data, leading to better performance. 
Finally, combining all results in the best performance, proving the importance of each component and design choice.

\boldparagraph{Ablations on Multi-Agent} In \tabref{tab:ablation_multi}, we analyze the effect of adaptive head with dynamic weights on multi-agent prediction. The results show that the performance is improved significantly with the adaptive head. %
This indicates that the adaptive head can adjust the weights according to the initial state of each agent. %

\section{Conclusion and Future Work}
We propose a novel efficient framework for predicting future trajectories in both multi-agent and single-agent settings where switching between the two requires only changing the endpoint prediction head. We propose dynamic weight learning to accurately predict the endpoints of multiple agents in the same scene reference frame. We demonstrate that our model reaches state-of-the-art performance in both single-agent and multi-agent settings without increasing the model size and consequently without sacrificing efficiency in inference time.

An interesting direction for future work might be incorporating stochastic latent variables into the endpoint prediction to improve the uncertainty in future predictions, \eg with separate latent variables for short-term and long-term goals. Another promising direction is learning the temporal dynamics of the scene to understand the relations better and improve efficiency without limiting assumptions of factorized attention. 
Like most of the existing work in trajectory forecasting, we assume the availability of an HD map where the past locations of agents are marked. The effect of imperfect perception on predicting future trajectories needs to be studied in future work to deploy these solutions. 

\clearpage
{\small
\bibliographystyle{ieee_fullname}
\bibliography{bibliography_long, egbib}
}

\clearpage

\twocolumn[{%
  \begin{@twocolumnfalse}
    \supptitle{Supplementary Material for \\
``ADAPT: Efficient Multi-Agent Trajectory Prediction with Adaptation"}
  \end{@twocolumnfalse}}]

\setcounter{section}{0}
\setcounter{table}{0}
\setcounter{figure}{0}

\begin{abstract}
In this supplementary, we present the details of implementation~(\secref{sec:imp_details}) and training~(\secref{sec:train_details}), perform a complexity analysis of our method compared to other methods (\secref{sec:complexity}), provide an additional ablation study on the model components (\secref{sec:quantitative}), identify failure cases of our method (\secref{sec:failure}), and finally visualize additional qualitative results (\secref{sec:qual}). We also visualize the result of multiple attention heads in \secref{sec:qual} specializing in different areas and agents in the scene, which might help future work on interpretability.
\end{abstract}
\section{Implementation Details}
\label{sec:imp_details}

\boldparagraph{Scene Representation} We follow VectorNet~\cite{Gao2020CVPR} in our polyline subgraph implementation to obtain the updated node features $\bv_i$ of the subgraph as follows:
\begin{align}
    \bv^{(l + 1)}_i =&~ ~\text{cat}\left(\text{MLP}_{l}(\bv^{(l)}_i), ~\text{pool}(\text{MLP}_{l}(\bv^{(l)}))\right) \nonumber
\end{align}
where $\bv_i^{(l)}$ denotes the feature vector of agent $i$ at layer $l$, pool the max-pool operation, and cat the concatenation operation. We use an MLP with 2 linear layers and ReLU for non-linearity with a layer normalization~\cite{Ba2016ARXIV} after the first layer. We set the layer number $l$ to $3$ and the size of the feature vector to $128$ for both the agent and the lane subgraph.

\boldparagraph{Interaction Modelling} For each multi-head attention block (MHAB), we set the number of attention heads to $8$ and apply a dropout rate of $0.1$ to the attention probabilities. We set the size of the hidden layer in the feed-forward networks to $128$ and the number of iterations $L$ to 3. 

\boldparagraph{Meta Info} Meta info includes the location of the agent at time $t$, $t - 1$, and the yaw angle at $t$. Locations are in 2D coordinates and the angle is in radians, resulting in a 5-dimensional vector. We concatenate the meta info to the corresponding agent feature before decoding.

\boldparagraph{Trajectory Predictor} For both dynamic and static heads, we use a 2-layer MLP with ReLU for the non-linearity and a layer normalization~\cite{Ba2016ARXIV} after the first layer. Differently from subgraphs, we use residual connections in the last layer. 
\section{Training Details}
\label{sec:train_details}
As mentioned in the paper, we use the variety loss to capture multi-modal futures by calculating the loss only for the most accurate trajectory over $K$ predicted ones. Given the ground truth trajectory $\{\bs_t\}_{t=1}^T$ and the predicted trajectory with the closest endpoint $\{\hat{\bs}_t\}_{t=1}^T$ for $T$ future steps, we train our model using the endpoint loss $\mathcal{L}_{end}$, the full trajectory loss $\mathcal{L}_{traj}$, and the trajectory classification loss $\mathcal{L}_{cls}$. $\mathcal{L}_{end}$ is the difference between the closest endpoint and the ground truth endpoint:
\begin{align}
    \mathcal{L}_{end} =&~ \mathcal{L}_{\text{Smooth-}\ell_1}(\hat{\bs}_T,~\bs_T)
\end{align}
where $\hat{\bs}_T$ is the endpoint of $\hat{\bs}$, \ie the prediction at time $T$. $\mathcal{L}_{traj}$ is the mean of the per-step difference between the predicted full trajectory, $\hat{\bs}$, and the ground truth trajectory, $\bs$:
\begin{align}
    \mathcal{L}_{traj} =&~ \frac{1}{T} \sum_{t=1}^{T} \mathcal{L}_{\text{Smooth-}\ell_1}(\hat{\bs}_t,~\bs_t)
\end{align}
Finally, $\mathcal{L}_{cls}$ is the Binary Cross Entropy Loss applied to the assigned probabilities $\bp$ of $K$ trajectories where the ground truth probability of the closest trajectory $\hat{\bs}$ is set to $1$ and the others to $0$:
\begin{align}
    \mathcal{L}_{cls} =&~ \mathcal{L}_{\text{BCE}}(\bp,~\by)
\end{align}
where $\by$ denotes the ground truth probabilities assigned. Overall, our loss is the sum of these three losses:
\begin{align}
    \mathcal{L} =&~ \mathcal{L}_{end} + \mathcal{L}_{traj} + \mathcal{L}_{cls}
\end{align}

\begin{figure}[t]
    \centering
    \begin{subfigure}[t]{0.25\textwidth}
       \includegraphics[width=0.95\linewidth, trim={0 2cm 26cm 0}, clip]{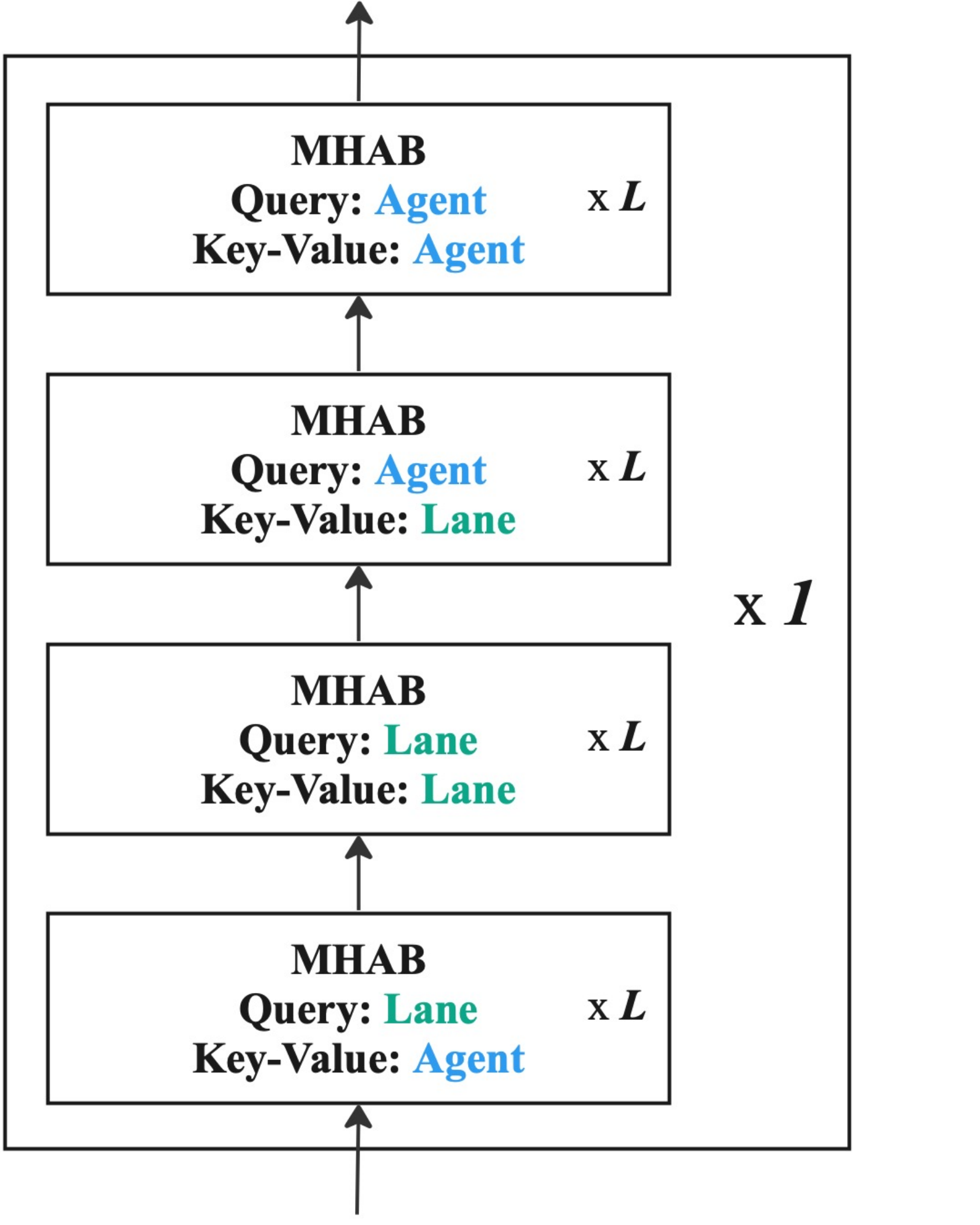}
       \caption{Sequential Order}
       \label{fig:sequential_attention} 
    \end{subfigure}%
    \begin{subfigure}[t]{0.25\textwidth}
        \includegraphics[width=0.95\linewidth, trim={26cm 2cm 0 0}, clip]{gfx/supp/attention_order.pdf}
        \caption{Iterative Order}
        \label{fig:iterative_attention} 
    \end{subfigure}
\caption{\textbf{The Order of Attention in Interaction.} Given a number of layers \textbf{\textit{L}}, there are two ways of applying attention to model interactions: sequential and iterative. In sequential order~\subref{fig:sequential_attention}, each type of interaction is considered  \textbf{\textit{L}} times sequentially. In iterative order~\subref{fig:iterative_attention}, each type of interaction is considered once in a single pass and the pass is repeated \textbf{\textit{L}} times.}
\label{fig:attention_orders}
\end{figure}

\section{Computational Complexity}
\label{sec:complexity}
In this section, we provide a comparison of the computational complexity according to the attention operations used in the existing approaches. We first define variables that define the number of elements. $N$, $M$, and $T$ correspond to the number of agents, lane elements, and time steps, respectively. $T$ can be decomposed into two variables, $T_p$ and $T_f$, which refer to past and future time steps, respectively. In general, the number of agents dominates the computation, then, the number of lanes followed by the fixed number of time steps, \eg $T=50$~($N>M>T$). While the number of lanes $M$ stays mostly uniform across scenes, the number of agents $N$ might vary significantly even for the same scene.

As addressed in the SceneTransformer~\cite{Ngiam2022ICLR}, directly applying attention to both time and agent axes results in high overhead, with the computational complexity of $\mathcal{O}((NT + M)^2)$ where $N$ is the number of agents, $M$ is the number of lane segments and $T$ is the number of time steps including both past and future. SceneTransformer reduces it to $\mathcal{O}(NT^2 + N^2T + NTM)$ with factorized attention over time and agent axes. 

Autobot~\cite{Girgis2022ICLR} does not include lane elements in their factorized attention steps. Contrary to SceneTransformer, their encoding and decoding phases consider only past and future time steps, respectively, resulting in the complexity of $\mathcal{O}(N{T_p}^2 + N^2{T_p} + N{T_f}^2 + N^2{T_f})$ where $T_p$ denotes the number of past time steps and $T_f$ denotes the number of future time steps.

HiVT~\cite{Zhou2022CVPR} does not use the standard multi-axis factorized attention but embraces a more efficient type of temporal interaction by considering only one agent for each time step and attending to only one feature over different time steps. Since HiVT follows an agent-centric approach and calculates agent features independently from each other, considering only one agent in their local scene does not result in information loss. However, the agent-centric approach comes with the overhead of $N$ runs of the same procedure. Considering scene normalization for each agent and global interaction in the end, HiVT has the overall complexity of
$\mathcal{O}(N^2{T_p} + N{T_p}^2 + NM)$. 

ADAPT has a clear advantage in terms of computational complexity over the existing approaches. Our computation is not bounded by $T$ as our subgraphs in vectorized encoder handle the temporal reasoning. Since we calculate the attention over only agents and lanes, ADAPT has the complexity of $\mathcal{O}(N^2 + NM + M^2)$ resulting from the attention operations in the interaction modeling. Removing the number of time steps $T$ out of the equation is the main reason behind the efficiency gain of ADAPT.

\begin{table}[b]
   \begin{center}
      \small
      \centering
      \def\arraystretch{1.2}
      \begin{tabular}{l | c c c}
         \toprule
          & $\text{mADE}_6$ & $\text{mFDE}_6$ & $\text{MR}_6$ \\
         \midrule
         w/o Iterative Att. & 0.673 & 0.971 & 0.086 \\
         w/o Dual Subgraph & 0.671 & 0.960 & 0.086 \\
         ADAPT & \textbf{0.668} & \textbf{0.948} & \textbf{0.083} \\
         \bottomrule
      \end{tabular}
   \end{center}
   \vspace{-0.2cm}
   \caption{\textbf{Single-Agent Ablation Study on Argoverse (Val.).} This table shows the effect of iterative attention and dual subgraph on the performance of single-agent prediction on the Argoverse validation set.}
   \label{tab:supp_ablation}
\end{table}

\section{Quantitative Results}
\label{sec:quantitative}

In this section, we present an additional ablation study to justify some minor design choices. Specifically, we investigated the effect of iterative \vs sequential order in interaction (\figref{fig:attention_orders}) and the effect of using two separate subgraphs for encoding agents and lanes. The results in \tabref{tab:supp_ablation} show that the iterative attention blocks outperform their sequential counterpart. This implies that updating intermediate features at each iteration, as opposed to the attention order used in LaneGCN~\cite{Liang2020ECCV}, leads to a better understanding of the relationship between agents and lanes. %
Furthermore, the use of separate polyline subgraphs for lanes and agents, which is in contrast to prior work~\cite{Gu2021ICCV, Gao2020CVPR}, produces better results. Overall, our decision choices on the architecture improve performance with better feature encoding.
\begin{figure*}[b]
\centering
    \begin{subfigure}[h]{0.47\textwidth}
        \includegraphics[width=1\linewidth, trim={3cm 1cm 3cm 1cm}, clip] {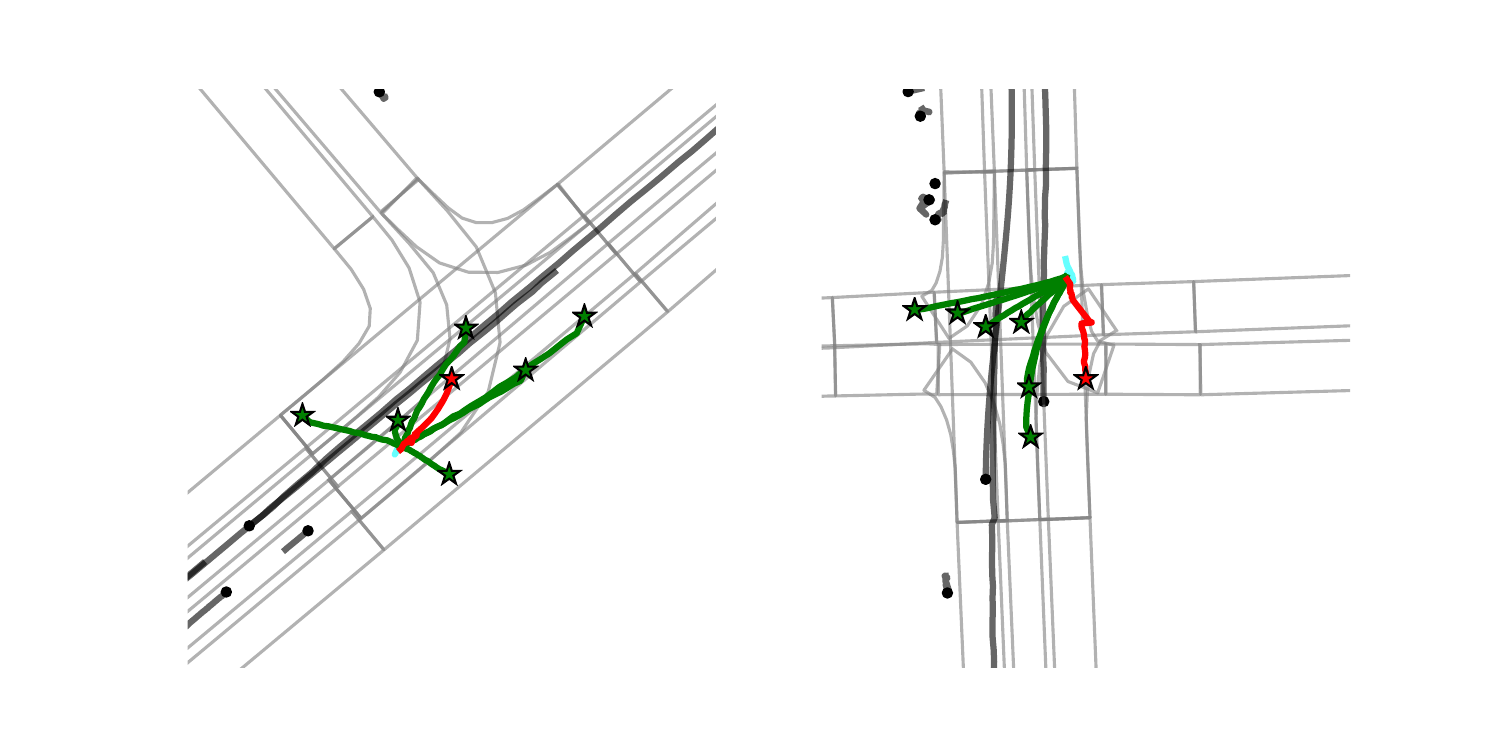}
        \caption{\textbf{Erroneous Input Trajectories.} The errors in the given trajectories, both past, and future, result in unreasonable futures due to inconsistent and uninformative past locations.}
        \label{fig:argo_fail_input_supp}
    \end{subfigure}%
    \hfill
    \begin{subfigure}[h]{0.47\textwidth}
      \includegraphics[width=1\linewidth, trim={3cm 1cm 3cm 1cm}, clip] {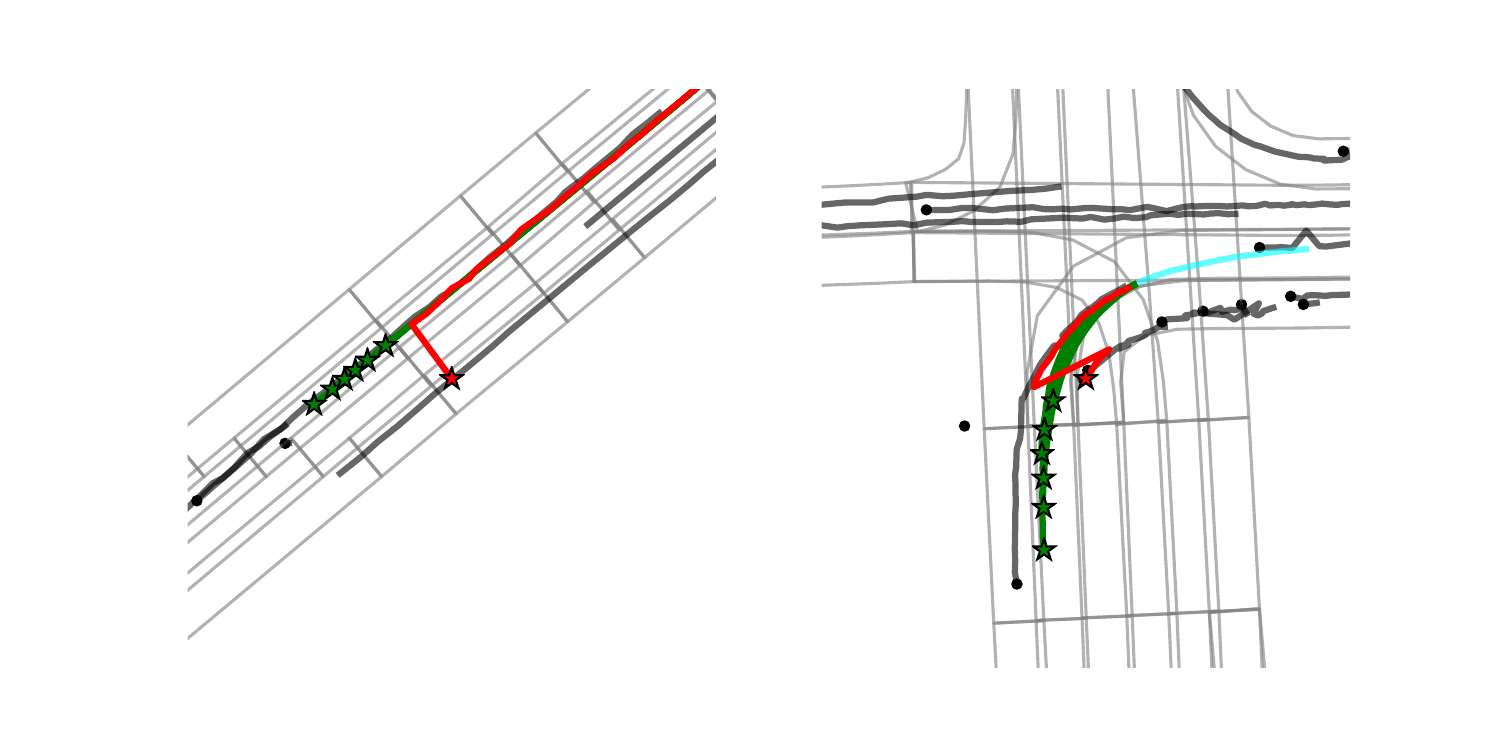}
    \caption{\textbf{Erroneous Ground Truth}. Erroneous (impossible) ground truth evaluates admissible trajectories as failures.}
    \label{fig:argo_fail_output_supp}
    \end{subfigure}
    
    \begin{subfigure}[h]{0.47\textwidth}
        \includegraphics[width=1\linewidth, trim={3cm 1cm 3cm 1cm}, clip] {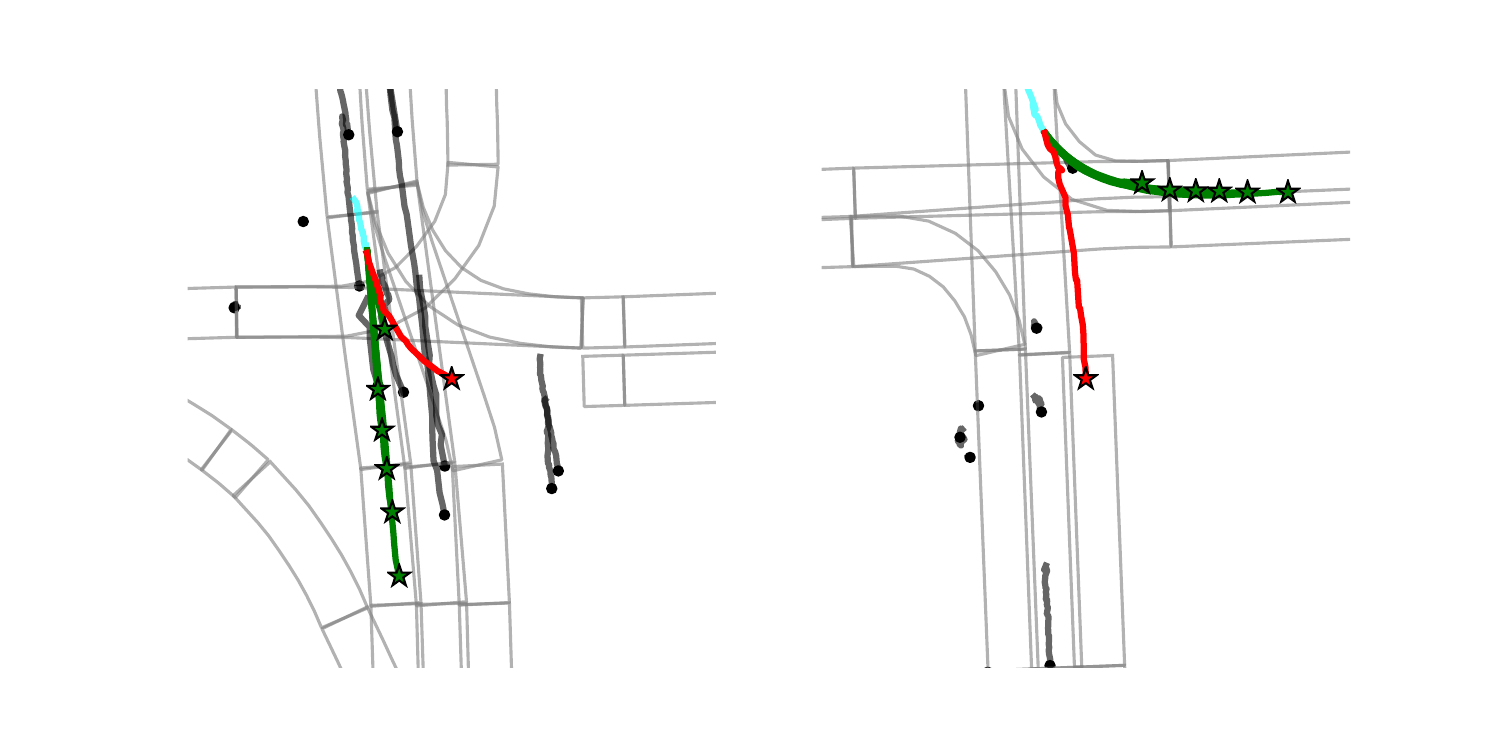}
        \caption{\textbf{Problems in the Input Map.} The missing lane information on the map causes the model to miss a possible future path.}
        \label{fig:argo_fail_lane_supp}
    \end{subfigure}%
    \hfill
    \begin{subfigure}[h]{0.47\textwidth}
        \includegraphics[width=1\linewidth, trim={3cm 1cm 3cm 1cm}, clip] {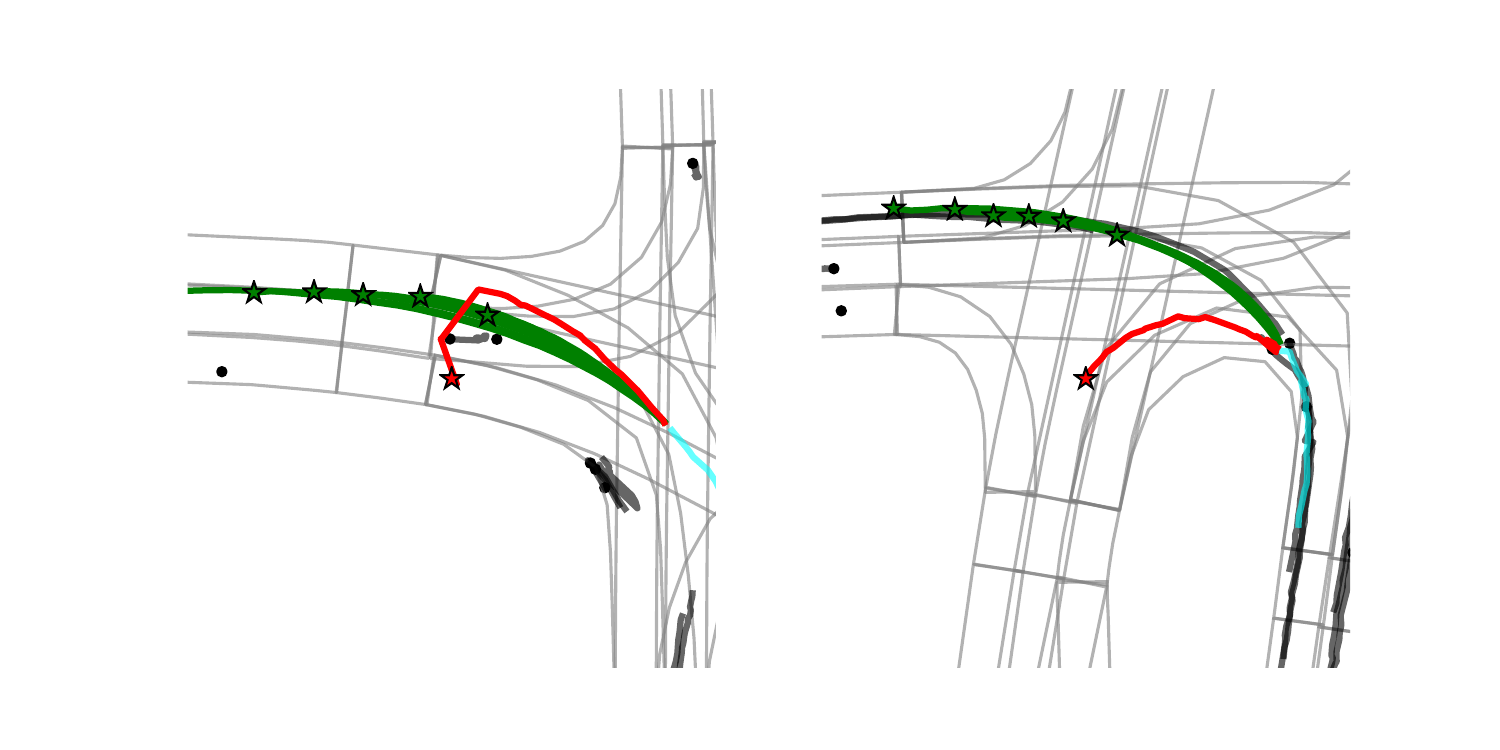}
        \caption{\textbf{Missing a Mode.} Rare behaviors cause the model to miss the mode corresponding to the ground truth.}
        \label{fig:argo_fail_behaviour_supp}
    \end{subfigure}
    
    \begin{subfigure}[h]{0.47\textwidth}
    \centering
        \includegraphics[width=1\linewidth, trim={3cm 1cm 3cm 1cm}, clip] {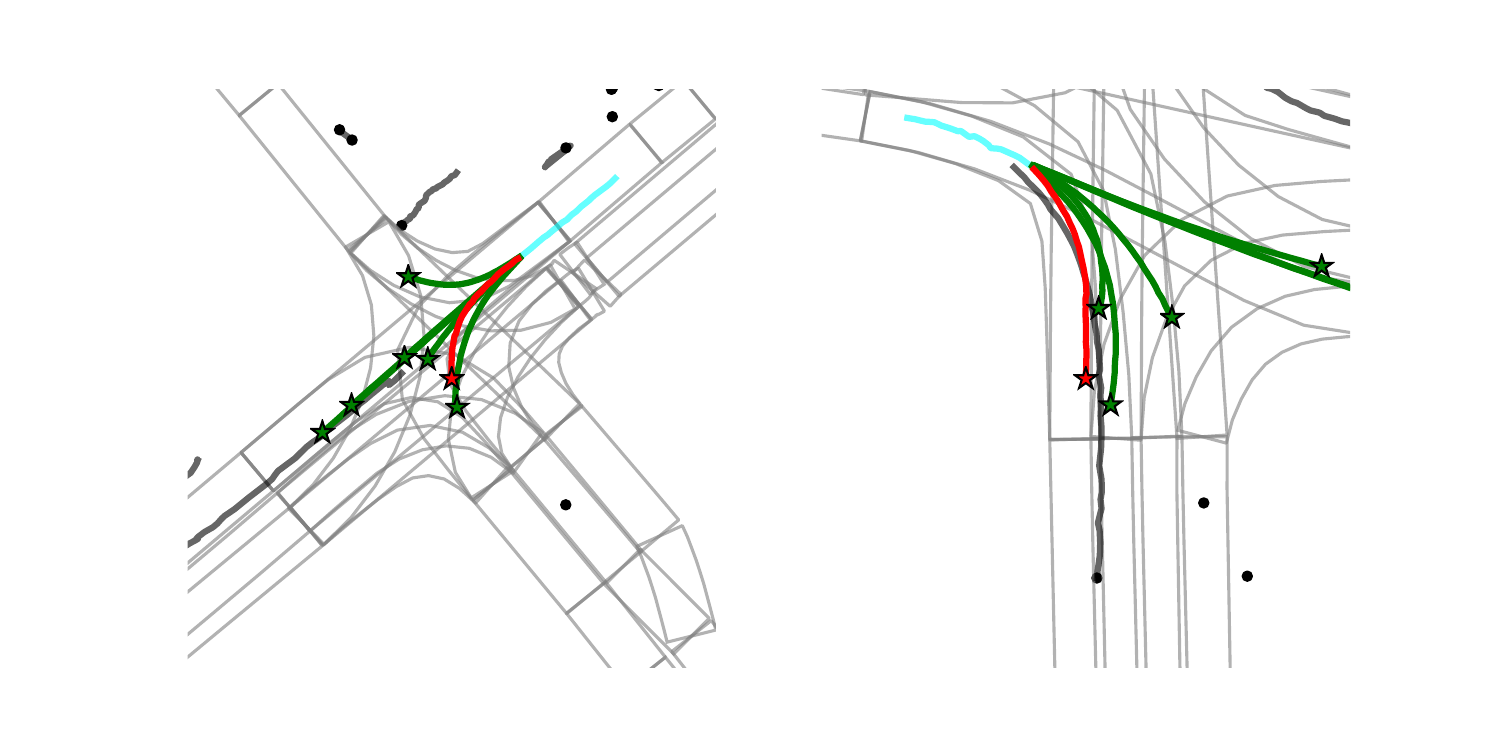}
        \caption{\textbf{Inaccurate Predictions}. The difference between the predicted and the ground truth endpoints is higher than the miss rate threshold, therefore this case is classified as a failure although ground truth intention is correctly captured by the predictions.}
        \label{fig:argo_fail_inaccurate_supp}
    \end{subfigure}

\caption{\textbf{Failure Cases on Argoverse}. We present some failure cases with their potential reasons.}
\label{fig:argo_fail_supp}
\end{figure*}

\section{Failure Cases}
\label{sec:failure}
In this section, we provide some failure cases and investigate possible reasons. We perform the analysis on single-agent predictions on Argoverse, since the miss rate in single-agent prediction is relatively higher than multi-agent predictions on Interaction. We identify three sources of error for failure cases: erroneous data, missing rare behaviors, and inaccurate predictions.

\boldparagraph{Erroneous Data} The accuracy of the provided input trajectories in the past directly affects the future predictions, since the future predictions are trained to be consistent with the past ones. Thus, defective or unstable history data causes incorrect future predictions as shown in \figref{fig:argo_fail_input_supp}. Moreover, defects in the future steps result in inaccurate evaluations of the predictions~(\figref{fig:argo_fail_output_supp}). Some problems such as id-switch and position-oscillation, resulting in unstable and incorrect ground truth future locations, are addressed in previous works as well~\cite{Ye2021CVPR, Song2021CoRL}.

Additionally, accurate map information plays an important role in future predictions because it directly affects the reasoning of the model about drivable areas. In the example shown in~\figref{fig:argo_fail_lane_supp}, predictions are intensified on a single mode \ie left turn, because of the missing lane that the ground truth trajectory follows.

\boldparagraph{Missing a Peculiar Mode} Despite the large number of scenarios on Argoverse, some behaviors are less frequently observed such as a u-turn or an abrupt lane change. These behaviors that are rare on the training set cause the model to miss the relevant mode at test time as shown in~\figref{fig:argo_fail_behaviour_supp}.

\boldparagraph{Precision of Predictions} Some predictions result in an error due to a lack of precision in the predicted trajectories despite correctly identifying intention. For example, in~\figref{fig:argo_fail_inaccurate_supp}, all possible paths are covered by the predictions but the difference between the closest endpoint and the ground truth endpoint is higher than the miss rate threshold.

\section{Qualitative Results}
\label{sec:qual}
\begin{figure*}[h!]
\centering
    \begin{subfigure}[h]{\textwidth}
      \includegraphics[width=1\linewidth, trim={5cm 0.5cm 5cm 0.5cm}, clip] {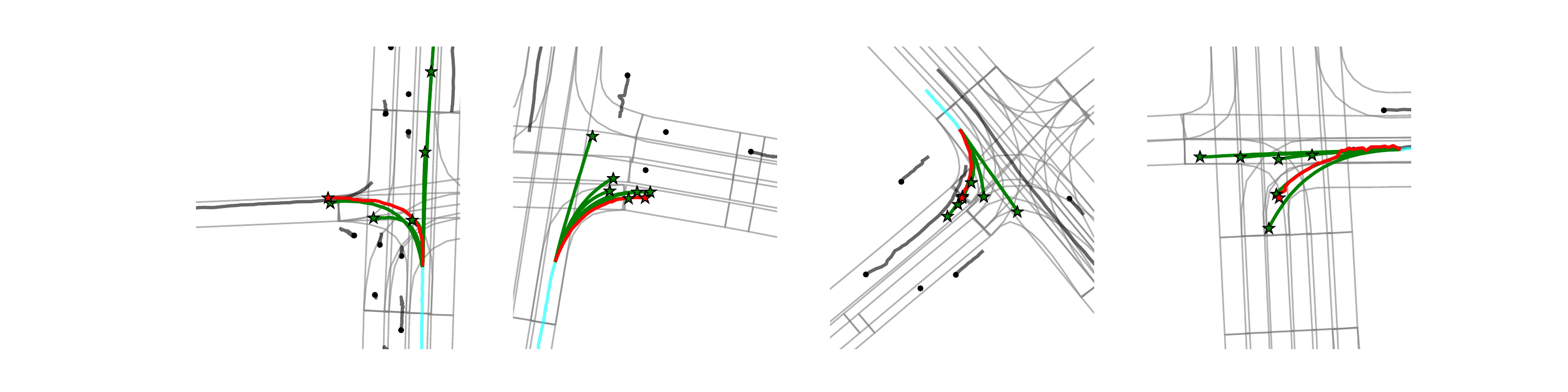}
    \end{subfigure}

    \begin{subfigure}[h]{\textwidth}
      \includegraphics[width=1\linewidth, trim={5cm 0.5cm 5cm 0.5cm}, clip]{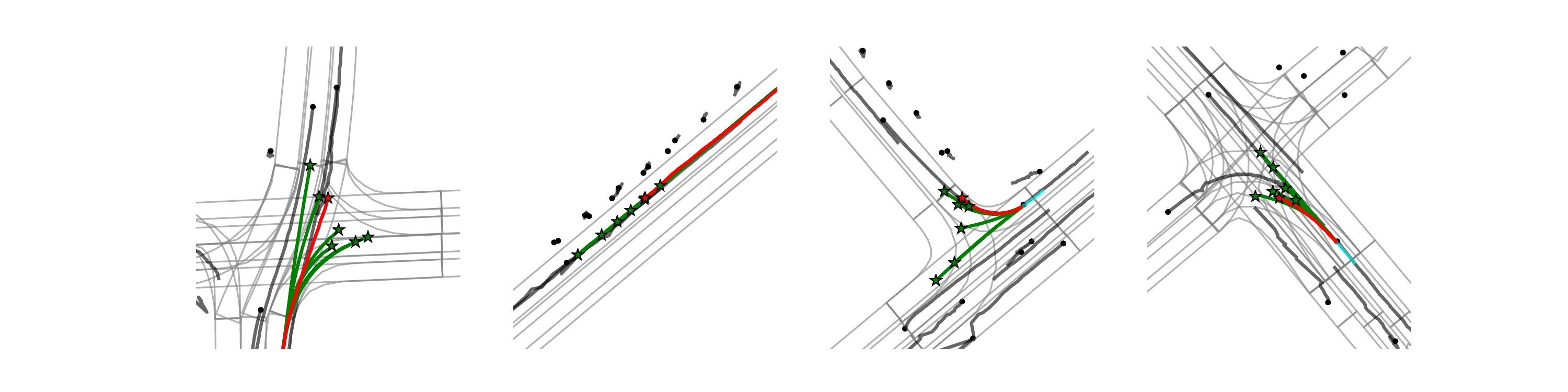}
    \end{subfigure} 
    
    \begin{subfigure}[h]{\textwidth}
      \includegraphics[width=1\linewidth, trim={5cm 0.5cm 5cm 0.5cm}, clip]{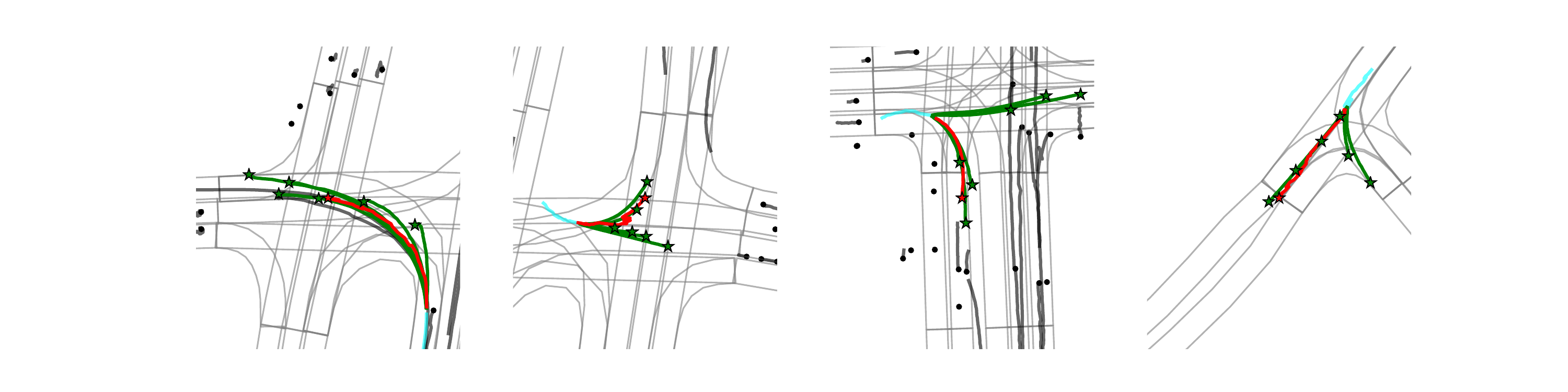}
    \end{subfigure} 
    
    \begin{subfigure}[h]{\textwidth}
      \includegraphics[width=1\linewidth, trim={5cm 0.5cm 5cm 0.5cm}, clip]{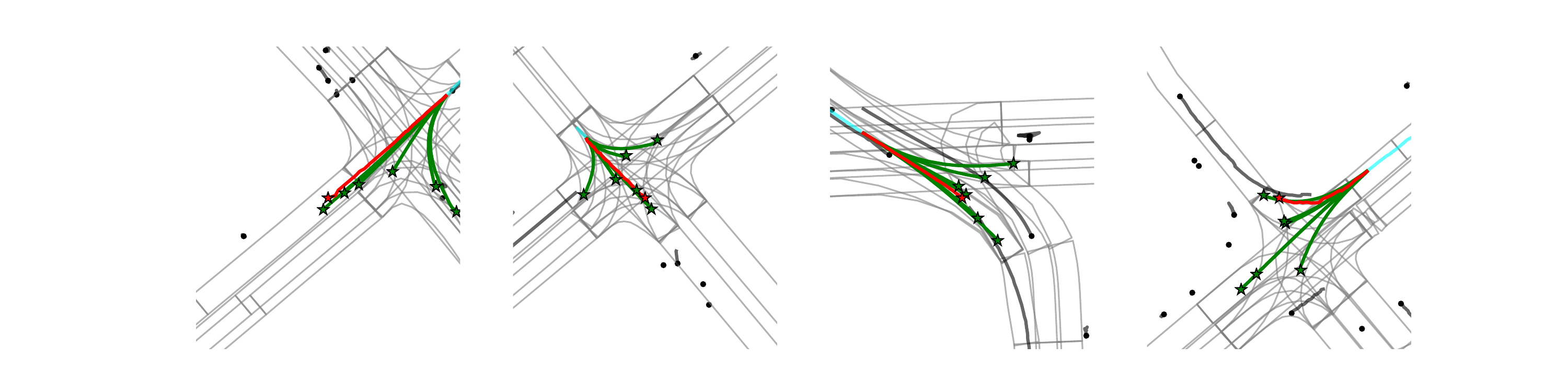}
    \end{subfigure} 
    
\caption{\textbf{Additional Qualitative Results for Single-Agent Predictions on Argoverse.} The red colored trajectory shows the ground truth future, the cyan shows the past trajectory of the agent of interest, and the green trajectories are the multiple predictions. Context agents are displayed in black. ADAPT can successfully predict a trajectory similar to the ground truth by also covering possible diverse trajectories for agent of interest.}
\label{fig:argo_qual_supp}
\end{figure*}

\begin{figure*}[h!]
\centering
    \begin{subfigure}[h]{\textwidth}
      \includegraphics[width=1\linewidth, trim={1.25cm 2cm 1.25cm 2cm}, clip] {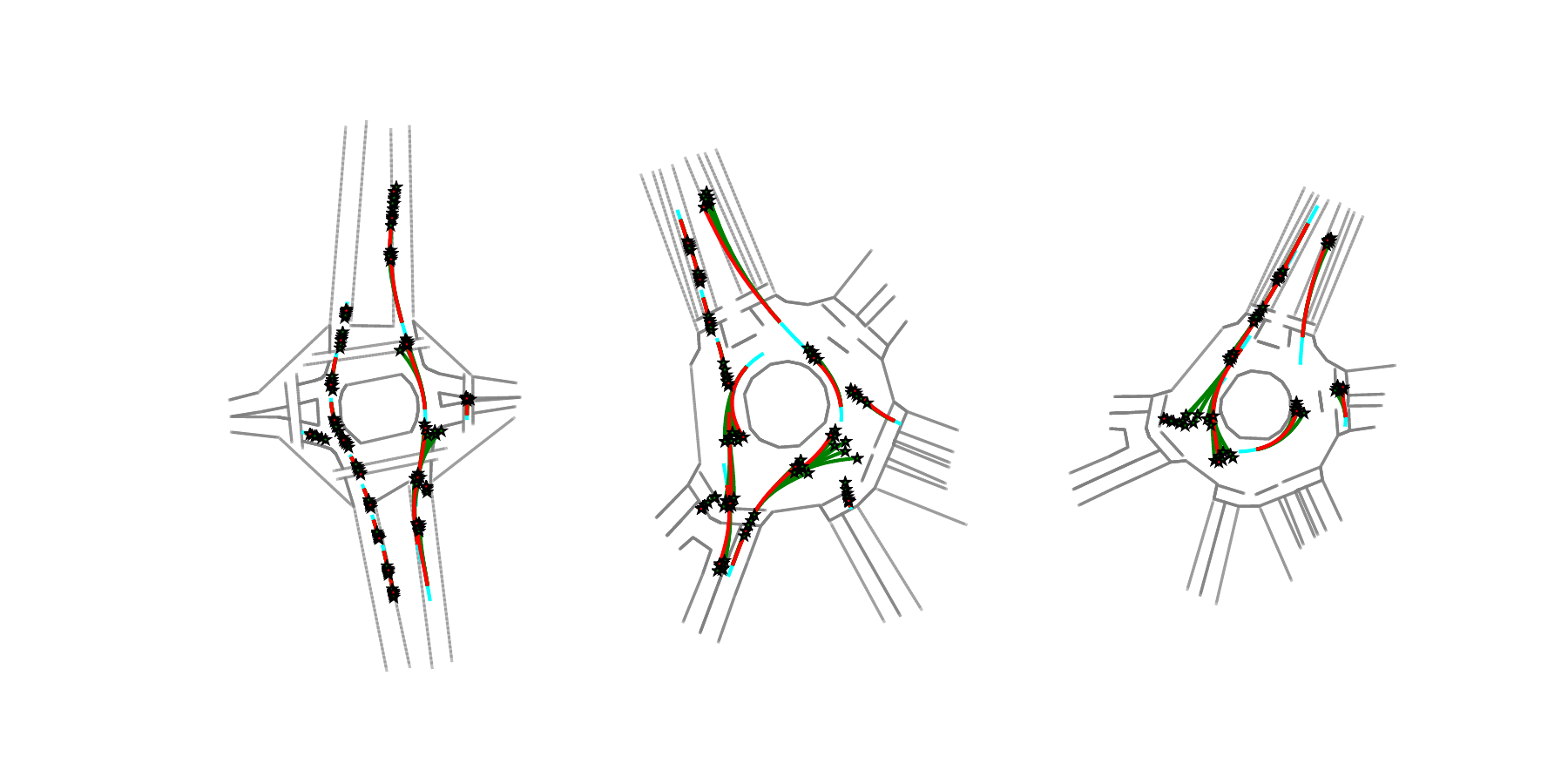}
    \end{subfigure}
    
    \begin{subfigure}[h]{\textwidth}
      \includegraphics[width=1\linewidth, trim={1.25cm 2cm 1.25cm 2cm}, clip] {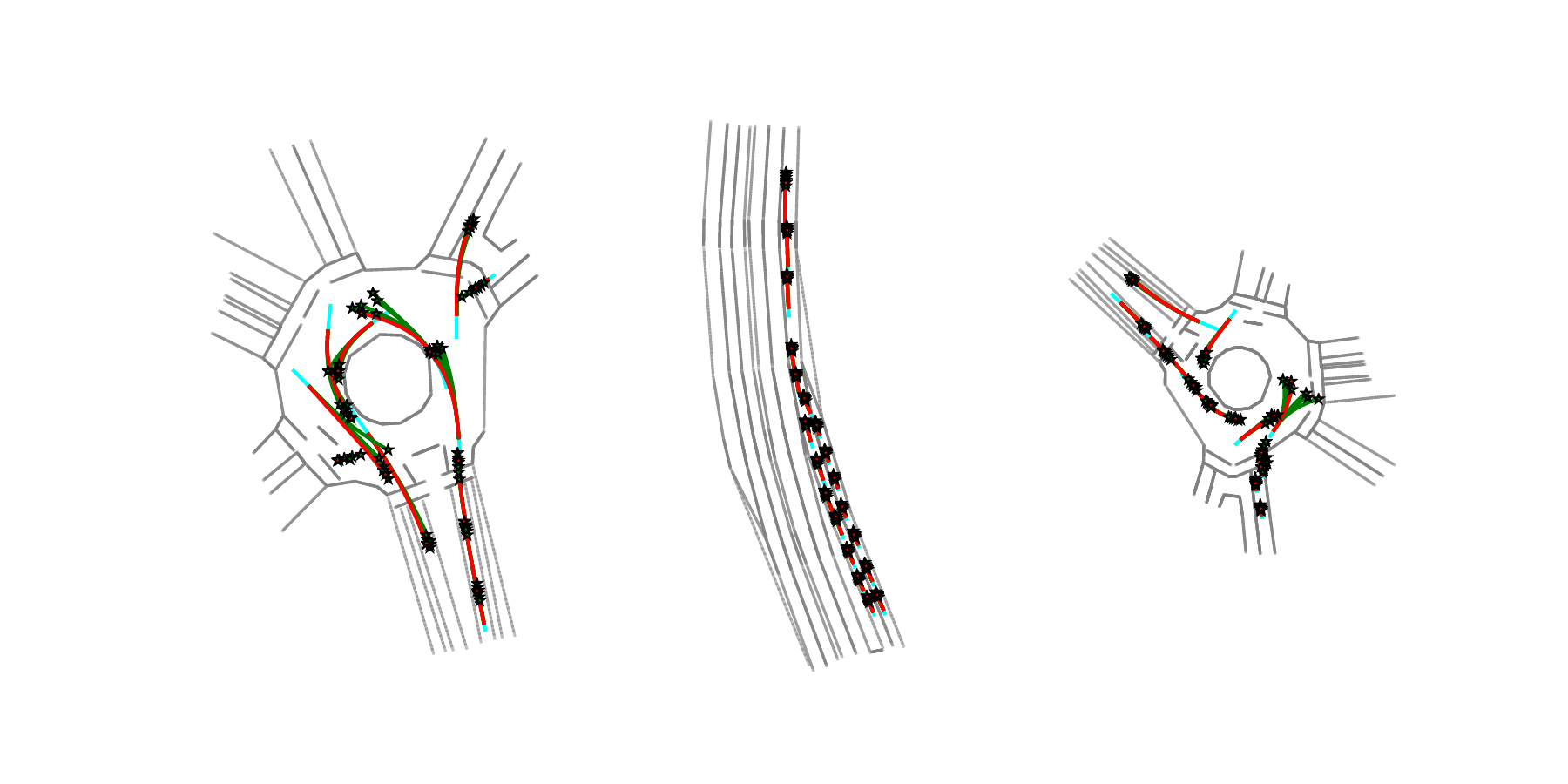}
    \end{subfigure}

    \begin{subfigure}[h]{\textwidth}
      \includegraphics[width=1\linewidth, trim={1.25cm 2cm 1.25cm 2cm}, clip] {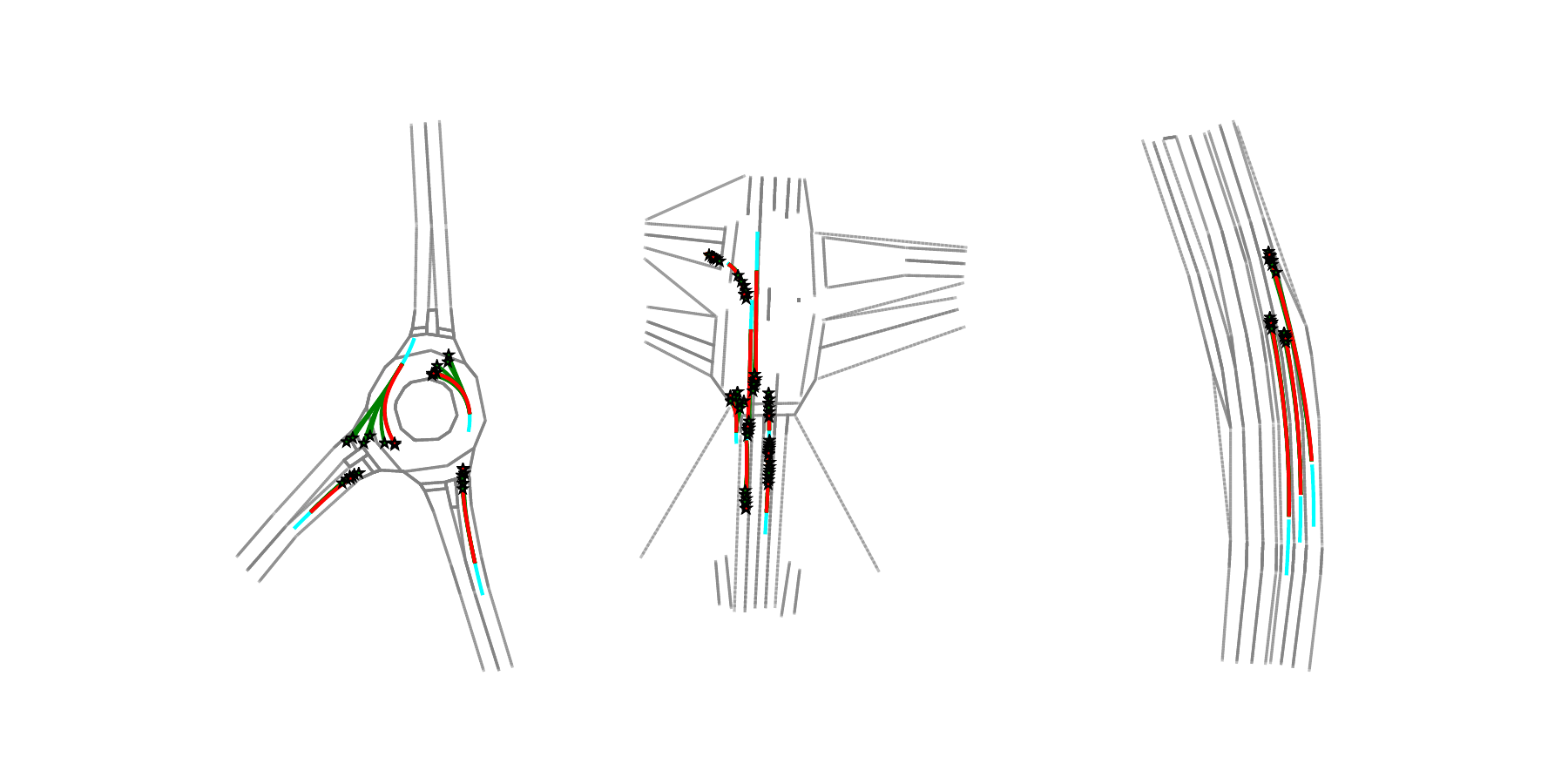}
    \end{subfigure}

\caption{\textbf{Additional Qualitative Results for Multi-Agent Predictions on Interaction.} The red colored trajectories show the ground truth future for each agent, the cyan shows the past trajectories of the agents, and the green trajectories are the predictions. ADAPT can successfully predict futures for each agent in a single forward pass without introducing additional overhead.}
\label{fig:interaction_qual_supp}
\end{figure*}

In this section, we provide additional qualitative results for both single-agent~(\figref{fig:argo_qual_supp}) and multi-agent~(\figref{fig:interaction_qual_supp}) predictions of ADAPT on Argoverse and Interaction validation sets, respectively.

\begin{figure*}[h!]
\centering
    \begin{subfigure}[h]{\textwidth}
      \includegraphics[width=1\linewidth, trim={1cm 0.5cm 1cm 1cm}, clip] {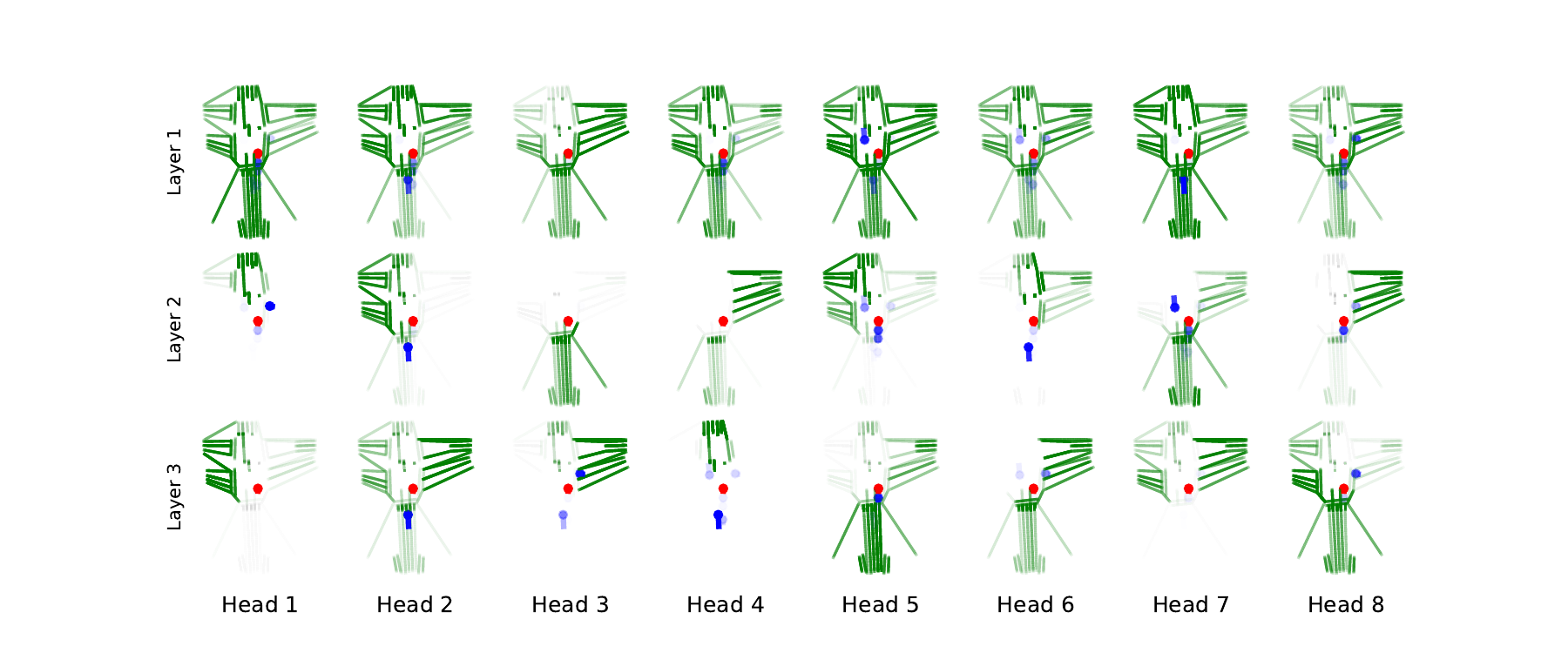}
    \end{subfigure}
    
    \begin{subfigure}[h]{\textwidth}
      \includegraphics[width=1\linewidth, trim={1cm 0.5cm 1cm 1cm}, clip] {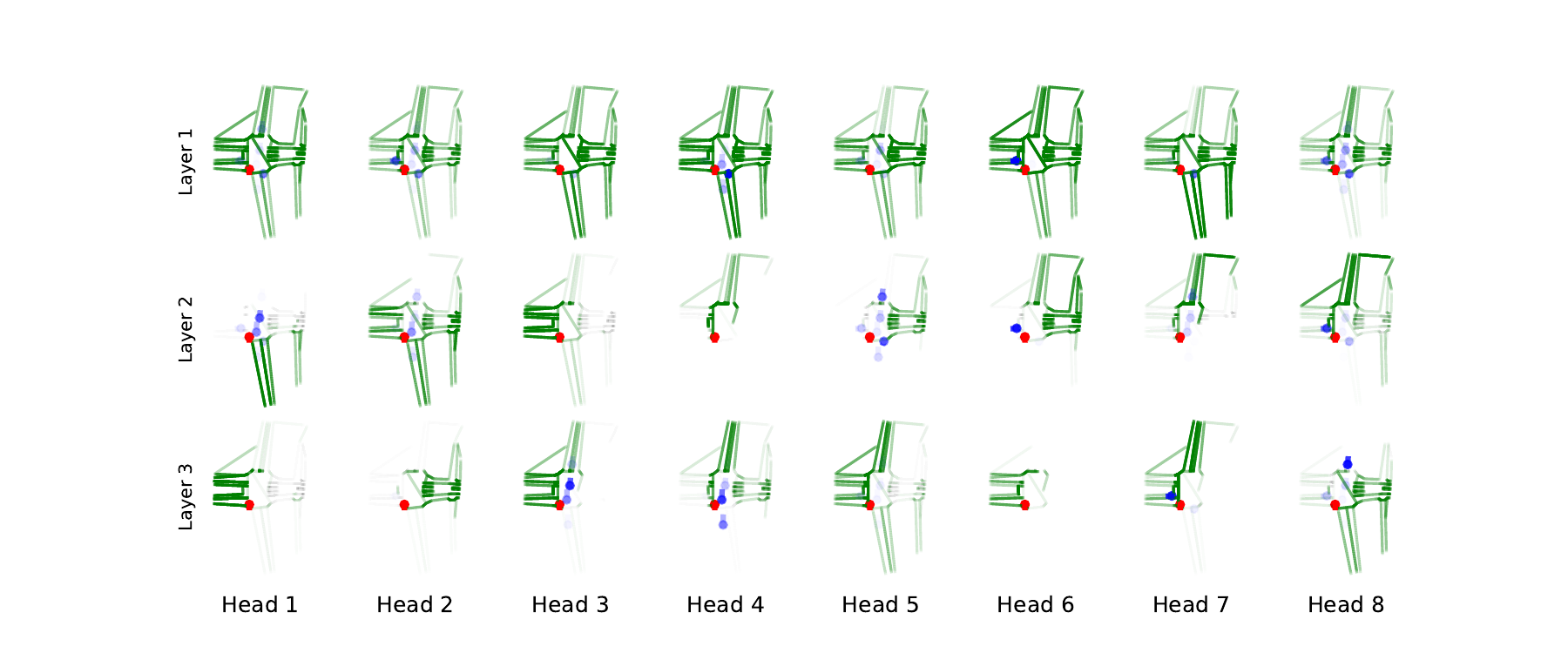}
    \end{subfigure}

    \begin{subfigure}[h]{\textwidth}
      \includegraphics[width=1\linewidth, trim={1cm 0.5cm 1cm 1cm}, clip] {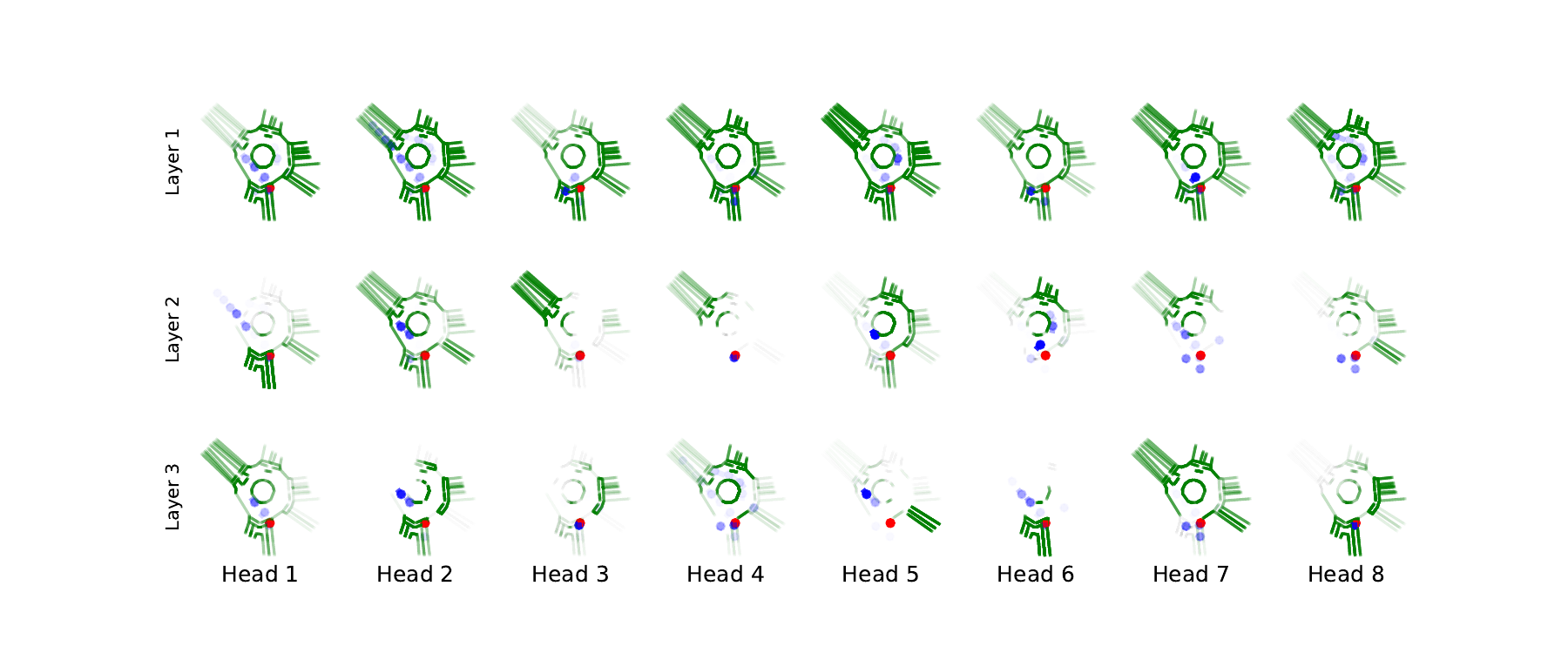}
    \end{subfigure}

\caption{\textbf{Visualization of Attention Scores on the Interaction}. We visualize multi-head attention from different layers for a selected agent (red). The attention probabilities for the agents (blue) and lanes (green) are the results of the Agent-Agent and the Lane-Agent modules, respectively. The transparency increases with lower attention probabilities. As the attention propagates towards higher layers, the attention heads specialize towards specific components such as lanes in the right turn, the vehicle in front, \etc.}
\label{fig:attention_supp}
\end{figure*}

\newpage
\boldparagraph{Focus of Attention Heads}
In~\figref{fig:attention_supp}, we visualize attention scores from different MHAB heads that are used to update an agent (red) in scenes from the validation split of the Interaction dataset. The attention scores are gathered from \textbf{AA} and \textbf{LA} modules for agents and lanes, respectively. In the first layer of interaction (the first row), attention heads do not focus on any specific scene elements yet as this is the first step where the agent is informed by the scene. As the agent has no prior information, attending to all scene elements without focusing on any is a reasonable choice in the first layer. On the other hand, in the next layer, each attention head specializes in some part of the map. For example, in the scene given in the upper set of rows, head 2 attends to lanes in the upper left of the map whereas head 4 attends to the upper right lanes. Some \textbf{AA} heads attend only to the agent itself and do not consider any other agents, \eg head 3 and head 4. In the last layer, heads still attend to some specific parts of the map and a subset of the agents. These visualizations confirm that different types of interactions are captured with the multi-head attention blocks of our model.

\end{document}